\documentclass{article} 
\usepackage{iclr2026_conference,times}

\usepackage{amsmath,amsfonts,bm}

\def\eqref#1{equation~\ref{#1}}

\def\1{\bm{1}}

\DeclareMathAlphabet{\mathsfit}{\encodingdefault}{\sfdefault}{m}{sl}
\SetMathAlphabet{\mathsfit}{bold}{\encodingdefault}{\sfdefault}{bx}{n}

\usepackage{hyperref}
\usepackage{url}
\usepackage{graphicx}
\usepackage{lipsum}
\newtheorem{observation}{Observation}
\usepackage{subcaption}
\usepackage{adjustbox}
\usepackage{multirow}
\usepackage{booktabs}
\usepackage{adjustbox}
\usepackage{wrapfig}
\usepackage{colortbl}
\usepackage{xcolor}
\usepackage{tcolorbox}
\usepackage{mdframed}
\usepackage{lipsum}

\title{G-KV:  Decoding-Time  KV Cache Eviction  with Global Attention}

\author{
Mengqi Liao$^{1,2}$ , Lu Wang$^{2}$, Chaoyun Zhang$^{2}$, Zekai Shen$^{1}$, Xiaowei Mao$^{1}$,\\
\textbf{Si Qin}$^{2}$, \textbf{Qingwei Lin}$^{2}$, \textbf{Saravan Rajmohan}$^{2}$, \textbf{Dongmei Zhang}$^{2}$ \& \textbf{Huaiyu Wan}$^{1,3}$\thanks{ Corresponding author.}\\
\\
$^1$School of Computer Science and Technology, Beijing Jiaotong University \\
$^2$MicroSoft  \\
$^3$Beijing Key Laboratory of Traffic Data Mining and Embodied Intelligence \\
\\
\texttt{mqliao@bjtu.edu.cn, wlu@microsoft.com, hywan@bjtu.edu.cn} 
}

\begin{document}

\iclrfinalcopy
\maketitle

\begin{abstract}

Recent reasoning large language models (LLMs) excel in complex tasks but encounter significant computational and memory challenges due to long sequence lengths. KV cache compression has emerged as an effective approach to greatly enhance the efficiency of reasoning. However, existing methods often focus on prompt compression or token eviction with local attention score, overlooking the long-term importance of tokens. We propose G-KV, a KV cache eviction method that employs a global scoring mechanism, combining local and historical attention scores to more accurately assess token importance. Additionally, we introduce post-training techniques, including reinforcement learning and distillation, to optimize models for compressed KV cache settings. The code of this paper is available on:
\url{https://github.com/microsoft/G-KV}.
\end{abstract}

\section{Introduction}

Large language models (LLMs) have garnered widespread attention and applications. Recently released reasoning models have demonstrated remarkable performance \citep{guo2025deepseek,team2025kimi, yang2025qwen3}, even in addressing complex tasks such as mathematics and coding. These reasoning models achieve significant improvements across various problems through long chain-of-thought (CoT) \citep{wei2022chain}, enabling iterative reflection and verification. 
However, the long CoT of reasoning models typically consists of thousands or even tens of thousands of tokens. This imposes a substantial increase in computational costs and KV cache memory consumption. Notably, the computation of attention becomes a critical bottleneck, as its complexity scales quadratically with the sequence length.

To overcome the bottlenecks of memory and computational complexity, numerous optimization methods for KV cache or attention mechanisms have been proposed \citep{li2024survey}. Among these, some methods prune the KV cache of tokens, significantly reducing computational overhead and memory consumption. However, most of these methods concentrate on the compression of the prompt's KV cache at the prefilling stage \citep{li2024snapkv, cai2024pyramidkv, feng2024ada, kim2025kvzip}. For reasoning tasks, where the output length often far exceeds the input length, limiting compression efforts to the prompt's KV cache results in only marginal benefits.

Although some methods support dynamically evicting tokens during the decoding stage \citep{song2025reasoning, cai2025r}, thereby maintaining consistently low KV cache requirements throughout the generation process, they rely solely on the attention scores of a few most recently generated tokens to determine which tokens to evict.  However, our experiments reveal that the importance of tokens can shift during the generation process. Such a localized perspective overlooks the long-term significance of tokens. In addition, the original models may fail to adapt to the  sparse attention patterns induced by KV cache compression, resulting in suboptimal performance. \cite{xiao2023efficient} and \cite{ chensepllm} train models with sparse attention through pre-training. However, the cost of pre-training is exceedingly high.

To address the limitations of the local attention, 
\textbf{(1)} we propose a \textbf{simple and effective global score}. This global score combines the local attention score with historical scores to assess the long-term importance of tokens, thereby avoiding the eviction of critical context that may reappear in future attention patterns. The global score can be seamlessly integrated with most existing methods and significantly enhances performance. 
Furthermore, \textbf{(2)} to enable the original model to adapt to the sparse attention pattern, we implement a \textbf{reinforcement learning framework specifically tailored for KV cache compression}, which eliminates the discrepancy between the training policy and the inference policy.
\textbf{(3)} Our experiments show that integrating the global score under a 512-token budget improves other methods by 5\% to 20\%. The RL framework we developed for KV cache compression is better suited for training models with compressed KV cache, achieving significantly superior performance compared to RL conducted directly on Full KV cache.

\section{Related Work}
KV cache compression methods can be broadly categorized into four classes \citep{li2024survey}: KV cache selection, merging \citep{kim2023compressed,nawrot2024dynamic,liu2024minicache}, quantization \citep{yao2022zeroquant,sheng2023flexgen,liu2024cachegen}, and low-rank decomposition \citep{chang2024palu,dong2024get}. KV cache selection is the most pertinent to our work.

\textcolor{black}{\textbf{Prefilling KV Cache Compression.}}  SnapKV \citep{li2024snapkv} and KVzip \citep{kim2025kvzip} determine which KV cache to retain by leveraging the attention score from an observation window or an appended specially designed prompts. PyramidInfer \citep{yang2024pyramidinfer} and PyramidKV \citep{cai2024pyramidkv} allocate varying KV cache budgets across different layers. Ada-KV \citep{feng2024ada} and HeadKV \citep{funot} propose allocating different budgets to individual attention heads. These methods predominantly focus on compressing the prompt. However, as the reasoning length continues to increase, merely compressing the prompt still faces computational and memory bottlenecks. 

\textcolor{black}{\textbf{Decoding-time KV cache Eviction.}} H2O \citep{zhang2023h2o} uses the accumulated attention received by each token as its score, but this can easily lead to the tail tokens being ignored in long sequences. MorphKV \citep{ghadiadialogue} and \cite{song2025reasoning} dynamically evicts tokens during decoding using the local attention score of the latest tokens, while R-KV \citep{cai2025r} introduces redundancy scores to further enhance the information density of the KV cache. Nevertheless, these methods are constrained to attention within local windows. Although CAKE \citep{qin2025cake} considers temporal attention shifts, it remains restricted to a local window. 

\section{Preliminary}
\label{sct:pre}
Dynamic token eviction methods \citep{ghadiadialogue, cai2025r,song2025reasoning} can be conceptualized within a unified framework. In this framework, the KV cache is compressed after every \( s \) tokens are generated. The most recent \( w \) generated tokens constitute the \textit{observation window}. The query states \( \mathbf{Q} \in \mathbb{R}^{h_\text{q} \times w \times d} \) of tokens in the observation window, alongside the cached key states \( \mathbf{K} \in \mathbb{R}^{h_\text{kv} \times l \times d} \) are employed to compute score using a function \( f(\mathbf Q, \mathbf K): \mathbb{R}^{h_\text{q} \times w \times d} \times \mathbb{R}^{h_\text{kv} \times l \times d} \rightarrow \mathbb{R}^{h_{\text{kv}} \times l} \).  Here, \( h_\text{q} \) and \( h_\text{kv} \) denote the number of heads for the query states and key states, respectively, while \( l \) represents the length of the KV cache.  For each head, the key states and value states corresponding to the top-\((b-w)\) scores are retained. Here, \(b\) denotes the budget size of the KV cache. Incorporating the KV cache from the observation window of length \( w \), the final compressed KV cache has a total length of \( b \).
\textbf{This framework effectively balances memory efficiency with the preservation of critical contextual information for future token generation. }

Typically, these methods involve computing the attention scores between the query states within the observation window and the cached key states. The $i$-th head attention score formula is as follows:

\begin{equation}
    \mathbf{A}_{[i,:,:]} = \text{softmax}\left(\frac{\mathbf{Q}_{[i,:,:]} \mathbf{K}_{[j,:,:]}^\top}{\sqrt{d}}\right) ,
\end{equation}

where \(\mathbf{A} \in \mathbb{R}^{h_\text{q} \times w \times l}\). Here, \(j\) represents the head index for the key states. For multi-head attention \citep{vaswani2017attention}, \(j = i\). However, in the case of multi-query \citep{shazeer2019fast} or group-query \citep{ainslie2023gqa} attention, \(j\) and \(i\) exhibit a one-to-many relationship.
To obtain the scores corresponding to each key state, a max-reduce operation is performed across the scores of multiple attention heads corresponding to each key head, resulting in \(\mathbf{A'} \in \mathbb{R}^{h_\text{kv} \times w \times l}\).
The final scores \(\mathbf{S}\) are then computed from \(\mathbf{A'}\) by applying a mean operation within the observation window,  
\begin{equation}
    \mathbf S_{i,j} = \frac{1}{w} \sum_{k=0}^{w-1} \mathbf A'_{i,k,j},
    \label{eq:local_score}
\end{equation} 
yielding \(\mathbf{S} \in \mathbb{R}^{h_\text{kv} \times l}\). This score reflects the significance of the key states and value states within the KV cache.

\section{Observation}
\label{sct:observation}

Dynamic token eviction methods are based on an intuitive assumption: tokens attended by the observation window are the most critical for subsequent generation. The strong performance of these methods suggests that this assumption holds some validity. However, \textbf{can a single observation window effectively determine which tokens are truly essential for subsequent generation?} To investigate this question, we devise the following experiment.

\begin{figure}[h]
    \centering
    \begin{subfigure}{0.14\linewidth}
        \centering
        \includegraphics[width=\linewidth, trim=340 140 430 100, clip]{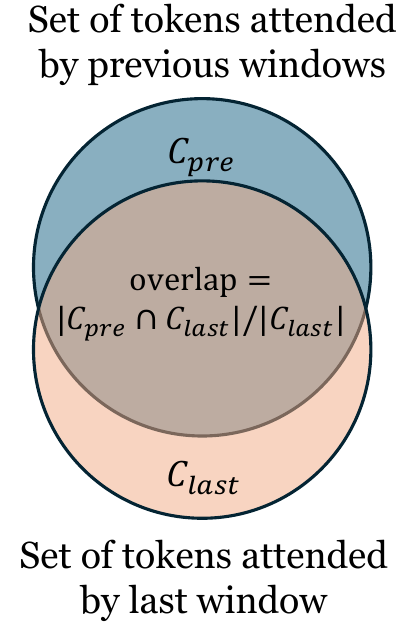}
    \end{subfigure}
    \begin{subfigure}{0.85\linewidth}
        \centering
        \includegraphics[width=\linewidth, trim=0 20 0 20, clip]{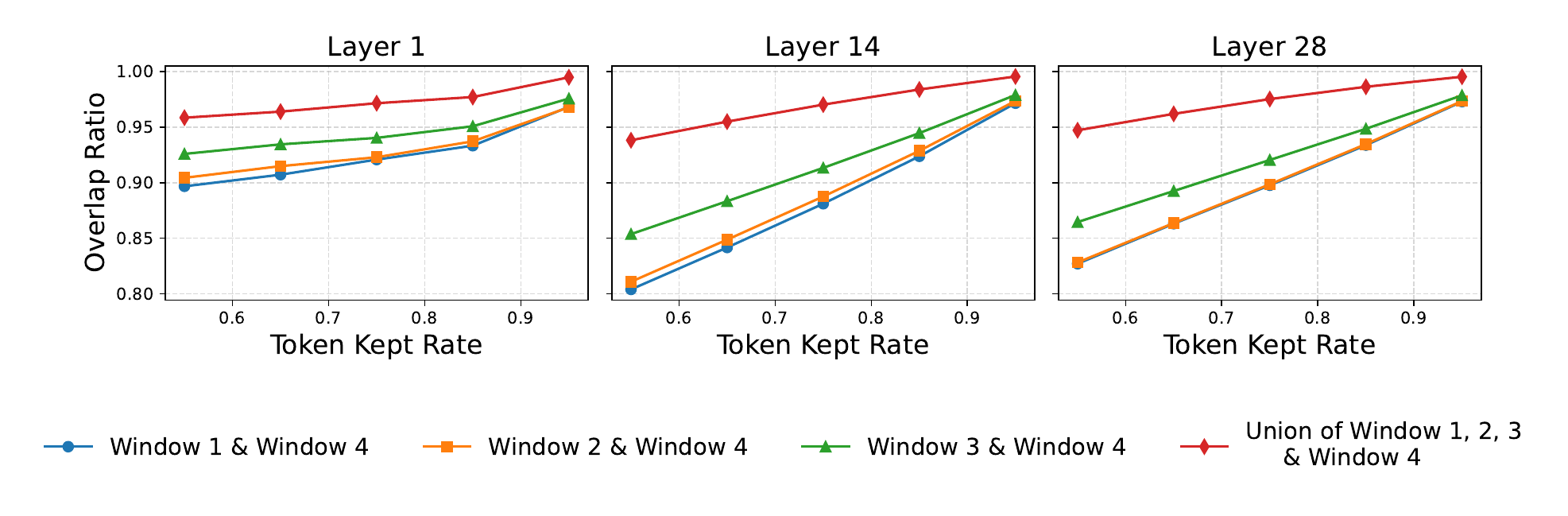}
    \end{subfigure}

    \caption{The left figure illustrates the calculation process for overlap. The right figure depicts the overlap between the last window and other windows across different layers. The horizontal axis represents the proportion of tokens retained.}
    \label{fig:win_consis_combined}
\end{figure}

We conducted inference using DeepSeek-Distill-Qwen-7B \citep{guo2025deepseek} on the AMC 2023 benchmark \citep{aops2023amc8}, performing 32 rollouts per question. The last 512 tokens of the generated output were divided into 4 observation windows, and for each window, the score \(\mathbf{S}\) (equation (\ref{eq:local_score})) was calculated based on the query status of tokens within the window and the key status of preceding tokens. Based on \(\mathbf{S}\), the \(p\%\) tokens with the highest scores were identified as the most attended tokens for each window. Subsequently, we quantified the \textit{overlap} between the token set attended to by the last observation window and those attended to by other windows, defined as the ratio of the intersection size to the size of the token set attended to by the last window.  The results, shown in Figure \ref{fig:win_consis_combined}, reveal that:

\begin{observation}
The tokens attended to by the last window are not fully consistent with those attended to by the earlier windows. As the number of retained tokens decreases, the inconsistency becomes more pronounced. 
\end{observation}

This finding demonstrates that the importance of tokens shifts across different windows. \textbf{If KV cache compression is performed based on scores computed from a single window, some tokens that possess long-term importance are likely to be prematurely evicted due to being temporarily overlooked by a single window.}

Furthermore, we computed the overlap between the last window and the union of all preceding windows (the \textcolor{red}{red}  line in Figure \ref{fig:win_consis_combined}). We find that:

\begin{observation}
The overlap between the token set attended to by the last window and the union of tokens attended to by all preceding windows is relatively higher. Notably, even when retaining only 55\% of the tokens, the overlap approaches 95\%.
\end{observation}

This observation further illustrates that the attention received by tokens is \textbf{intermittent}. On the other hand, it also indicates that \textbf{tokens receiving significant attention are highly likely to have been similarly attended to by at least one preceding observation window.}

\section{Training-Free KV Cache Compression with Global Attention}

As previously discussed, scores computed from a single window are insufficient to effectively capture the long-term importance of tokens. To address this limitation, we aim to determine which tokens should be evicted by leveraging their attention scores across a broader context.  

\textcolor{black}{The characteristics of human memory reveal that memories revisited multiple times become increasingly reinforced, whereas those left unreviewed for extended periods gradually diminish.
Inspired by this, we propose a global score to quantify the degree to which tokens are attended to throughout the decoding process. We introduce a \textbf{memory decay rate}, $\alpha \in [0,1]$, to promote the eviction of tokens that no longer attract attention.
We experiment with three different forms for calculating the global score: \textit{max}, \textit{average}, and \textit{summation}. For \(i < b-w\), the three different forms of global scores are calculated using the following formulas:}

\begin{equation}
\textcolor{black}{
 \mathbf{F}_t[:, i] = \max \left(\alpha \cdot  \mathbf{F}_{t-1}[:, i] , \frac{\mathbf{S}_{t}[ :,i]}{\max_j(\mathbf{S}_{t}[ :,j])}\right),
 }
\label{eq:main}
\end{equation}

\begin{equation}
\textcolor{black}{
 \mathbf{F}_t[:, i] = 
\alpha \cdot  \mathbf{F}_{t-1}[:, i] +(1-\alpha)\cdot \frac{\mathbf{S}_{t}[ :,i]}{\max_j(\mathbf{S}_{t}[ :,j])},}
\label{eq:main}
\end{equation}

\begin{equation}
\textcolor{black}{
 \mathbf{F}_t[:, i] = 
\alpha \cdot  \mathbf{F}_{t-1}[:, i] + \frac{\mathbf{S}_{t}[ :,i]}{\max_j(\mathbf{S}_{t}[ :,j])},}
\label{eq:main}
\end{equation}

Here, \(\mathbf{F}_{t-1} \in \mathbb{R}^{h_\text{kv} \times (b-w)}\) represents the historical global scores from the previous step, while \(\mathbf{F}_t \in \mathbb{R}^{h_\text{kv} \times l}\) denotes the global scores in the current step. The attention scores $\mathbf S_t$ (equation (\ref{eq:local_score})) are normalized by the maximum values within each attention head. \textcolor{black}{Since only \(b-w\) tokens have scores from the previous step, for \(i \geq b-w\), \( \mathbf F_t = \frac{\mathbf{S}_{t}[:,i]}{\max_j(\mathbf{S}_{t}[:,j])}\).} Based on \(\mathbf{F}_t\), we select \(b-w\) tokens whose corresponding KV cache is retained, and the \(\mathbf{F}_t\) values of the retained tokens are subsequently recorded for use in the next compression step. At the first compression step, as \(\mathbf{F}_{t-1}\) is not available, KV cache selection is performed directly based on \(\mathbf{S}_t\).

\begin{figure}[h]
    \centering
    \includegraphics[width=0.95\linewidth, trim=0 110 20 70, clip]{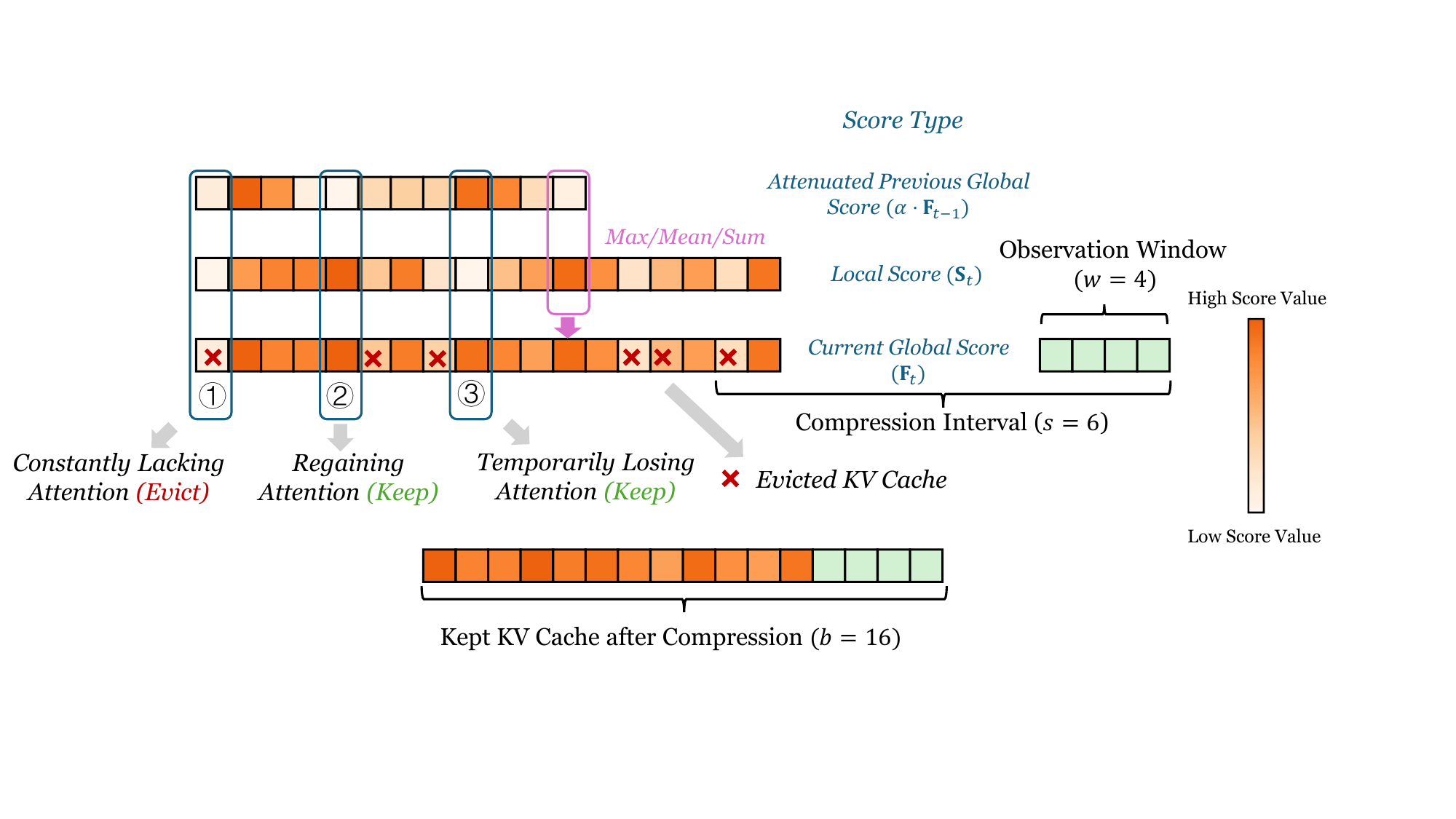}
    \caption{This figure illustrates the computation process of the global score. Each block represents the KV cache of a token, with the block's color indicating its score (darker color represents higher scores).}
    \label{fig:main}
\end{figure}

The score \(\mathbf{F}_t\) takes into account both the attention received in the current observation window and the attention from preceding windows, and we refer to it as the \textbf{global score}. In contrast, \(\mathbf{S}_t\), computed within a single window, is referred to as the \textbf{local score}. Figure \ref{fig:main} illustrates several scenarios that emerge when utilizing the global score:
\begin{itemize}
    \item \textbf{Low \(\mathbf F_{t-1}[i,j]\) and low \( \mathbf S_t[i,j]\):} This implies that the \(i\)-th KV head's \(j\)-th token consistently receives very little attention across multiple consecutive windows. Such tokens are considered insignificant and are therefore eligible for eviction.
    \item \textbf{Low \(\mathbf F_{t-1}[i,j]\) and high \( \mathbf S_t[i,j]\): } This suggests that the token temporarily lost attention in the previous observation window but regained attention in the current window. These tokens are re-engaged in the ongoing context.
    \item \textbf{High \(\mathbf F_{t-1}[i,j]\) and low \(\mathbf S_t[i,j]\): } This signifies that the token, while not receiving attention in the current window, was highly attended to in previous windows. Unlike other methods that might immediately evict such tokens, we choose to retain them because these tokens are highly likely to be attended to again in the future.
\end{itemize}

Compared to the local score, the global score better reflects the long-term importance of a token.  In addition, our analysis reveals that even with the KV cache compression algorithm, the attention scores remain highly sparse, as detailed in Appendix \ref{sct:sparse_nature}. Each observation window focuses on only a small subset of tokens within the compressed KV cache. \textbf{By employing global scores, the compression algorithm retains a small subset of tokens that are highly attended to by each observation window. Furthermore, tokens that are likely to receive significant attention in future windows are highly likely to be included within the union of these subsets.}


\section{Enhancing KV Cache Compression through Training}


\begin{wrapfigure}{r}{0.5\linewidth}
    \centering
    \vspace{-1em} 
    \includegraphics[width=\linewidth, trim=90 40 80 60, clip]{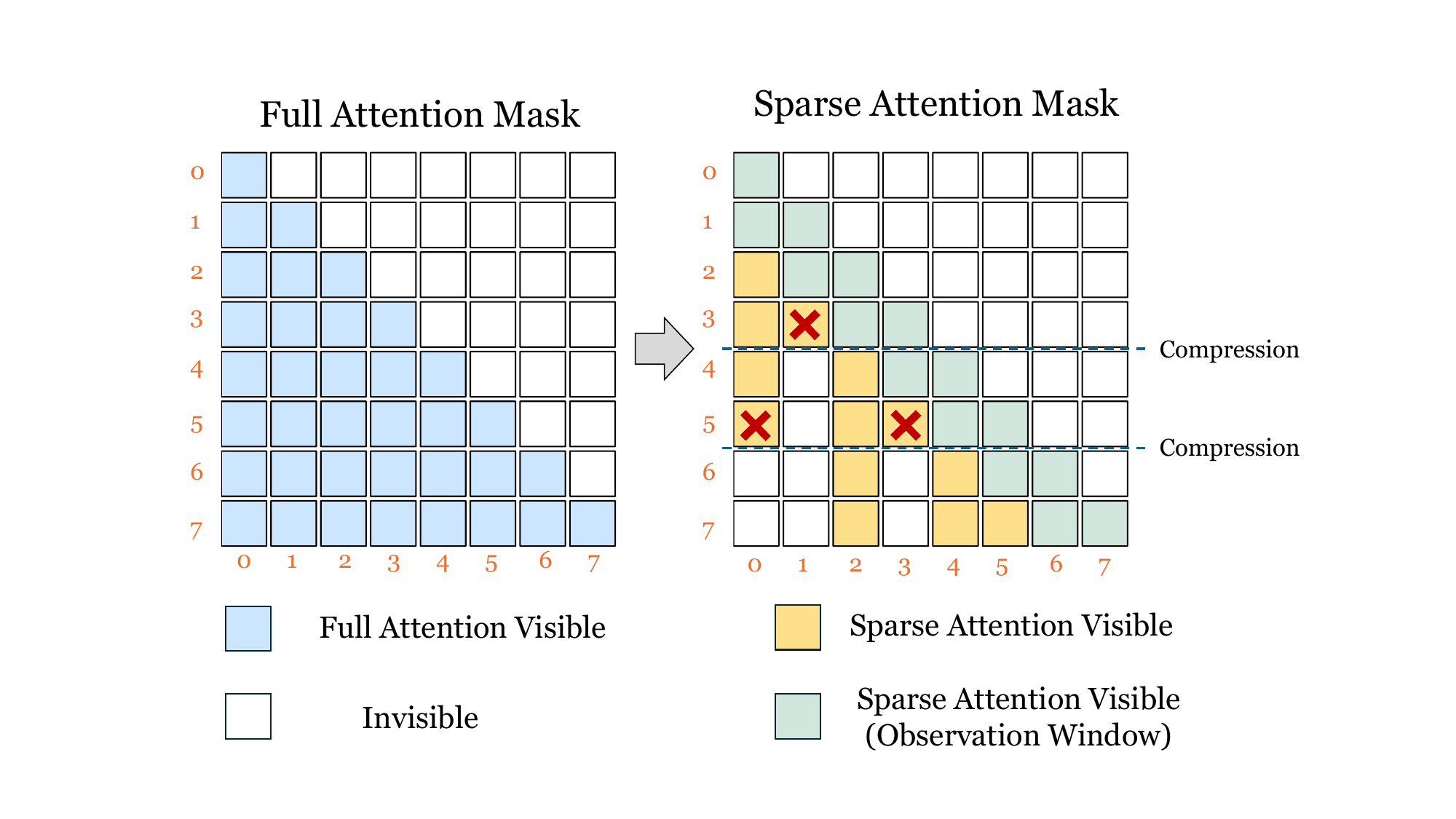}
    \caption{Illustration of the sparse attention mask. If the block at the \(i\)-th row and \(j\)-th column is visible, it indicates that the \(j\)-th token in the sequence can attend to the \(i\)-th token. Red crosses represent tokens evicted during a KV cache compression process; these tokens are invisible to newly generated tokens in subsequent steps.}
    \label{fig:sparse_attn}
    \vspace{-1em} 
\end{wrapfigure}

Dynamic token eviction can be seen as a form of sparse attention, where the KV cache of evicted tokens becomes inaccessible to subsequent tokens. Figure \ref{fig:sparse_attn} shows the sparse attention mask corresponding to the KV cache compression process. We define the original policy (LLMs) as $\pi_\theta$, with $\theta$ representing model parameters, and the policy with KV cache compression or sparse attention as $\pi'_\theta$. 
The original model $\pi_\theta$ is trained with full attention, relying on complete context. After compression, the policy $\pi'_\theta$ operates in a constrained context environment with unchanged parameters $\theta$, making $\pi'_\theta$ sub-optimal in this setting.


We aim to enable the model to adapt to the condition of KV cache compression. We explore post-training methods for this purpose. Specifically, we implemented a reinforcement learning (RL) framework that \textbf{supports generation with KV cache compression and training with sparse attention masks}. In this framework, sampling is performed directly by the policy $\pi'_\theta$. During generation, the positions of the tokens actually evicted are recorded and used to construct the sparse attention mask. The optimization objective is as follows:  

\begin{align}
    \mathcal{J}(\theta) = &\mathbb{E}_{ \{y_i\}_{i=1}^G \sim \pi'_{\theta_{\text{old}}}(\cdot | x)} \Bigg[
\frac{1}{G} \sum_{i=1}^G \frac{1}{|y_i|} \sum_{t=1}^{|y_i|}
\textcolor{black}{\min \Big( r_{i, t}(\theta) \hat{A}_{i, t}, \text{clip}(r_{i, t}(\theta), 1 - \epsilon, 1 + \epsilon) \hat{A}_{i, t} \Big)}
\Bigg],
\end{align}
where $ r_{i, t}(\theta) = \frac{\pi'_\theta(y_{i, t} \mid x, y_{i, <t})}{\pi'_{\theta_{\text{old}}}(y_{i, t} \mid x, y_{i, <t})}
$  and  $\hat{A}_{i, t} = \frac{r_i - \text{mean}(\{r_j\}_{j=1}^G)}{\text{std}(\{r_j\}_{j=1}^G)}$.
Here, \(r_i\) represents the reward associated with the response $y_i$. For each input $x$, we perform $G$ times sampling, where $y_{i, t}$ denotes the $t$-th token in the output of the $i$-th sampling. 
\textcolor{black}{This is the optimization objective of GRPO \citep{shao2024deepseekmath} without KL regularization.}
Moreover, training on outputs truncated due to the maximum output length constraint may introduce interference \citep{yu2025dapo}. To address this, we directly set their advantages to zero. Nevertheless, this RL method may only be suitable for tasks where the rewards of outputs can be easily verified. Consequently, we propose a more general distillation-like method, as detailed in Appendix \ref{sct:distill}.

\section{Experiment}

\subsection{Experiment Setup}

\textbf{Benchmark and Dataset}. \textcolor{black}{We evaluate the model's reasoning capabilities in the domains of mathematics and coding.} For the mathematics domain, we employ AMC 2023 \citep{aops2023amc8} and AIME 2024 \citep{aops2024aimei} as benchmarks. AMC is designed for middle-school students as an entry-level mathematical competition, while AIME serves as a critical gateway to advanced mathematics contests, featuring more challenging problems.  \textcolor{black}{For the coding domain, we conduct evaluations on LiveCodeBench \citep{jain2024livecodebench}, which includes programming competition problems of varying difficulty levels.}
For RL training, we use the DeepScaleR-40k \citep{deepscaler2025} dataset, which incorporates mathematical problems of varying difficulty levels from different datasets. Additionally, 27k correct reasoning-based responses are sampled from the DeepScaleR-40k dataset using DeepSeek-R1-Distill-Qwen-7B, and these samples are utilized for distillation. 

\textbf{Model}. We evaluate our approach using DeepSeek-R1-Distill-Qwen-7B and DeepSeek-R1-Distill-Llama-8B \citep{guo2025deepseek}. These models are reasoning models distilled from DeepSeek-R1 using Qwen 2.5 \citep{team2024qwen2} and LLaMA 3.1 \citep{grattafiori2024llama}, respectively.

\textbf{Evaluation Protocol and Metrics}. For sampling, we set the temperature to 0.6 and the top-p parameter to 0.95. Unless otherwise specified, for the AMC 2023, the maximum sequence length is configured to 16k, while for the AIME 2024, it is set to 32k.
We use pass@1 \citep{chen2021evaluating} as our evaluation metric, which is an unbiased estimate of the probability that the model answers a question correctly on the first attempt. For each question, we perform sampling 32 times to estimate the pass@1 score.

\subsection{\textcolor{black}{The Ablation and Comparison of Global Score}}
\label{sec:replace}

\textcolor{black}{In this section, we conduct experiments on different forms of global scores and various values of \(\alpha\), comparing them with the Local Score. Additionally, \textbf{CAKE} \citep{qin2025cake} uses the attention variance within a local window to represent the degree of attention fluctuation, which we refer to as the attention shift score, and we compare our method against it.}

\textbf{Implementation Details}. We set the KV cache budget to 512 ($b=512$), the observation window size to 16 ($w=16$), and perform compression after generating every 128 tokens ($s=128$).

\begin{figure}[h]
    \centering
    \includegraphics[width=0.9\linewidth]{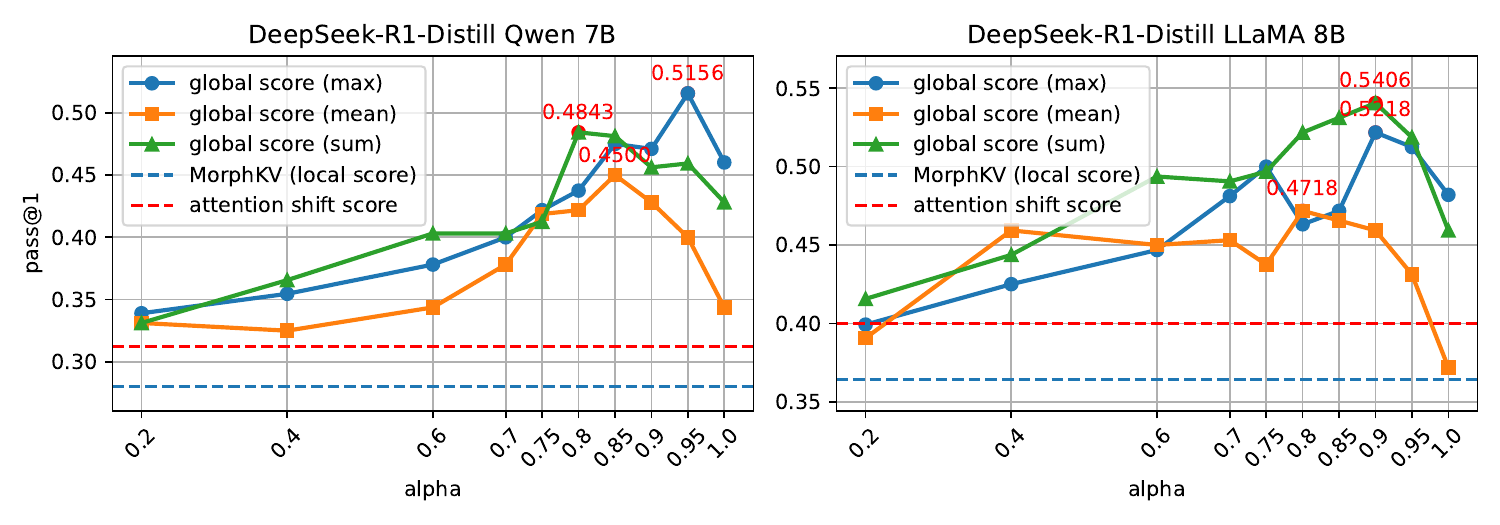}
    \caption{\textcolor{black}{Performance of different methods on the AMC 23 benchmark. }}
    \label{fig:global_exp}
\end{figure}

\textbf{Analysis}.  \textcolor{black}{As shown in Figure \ref{fig:global_exp}, all three forms of global score deliver significant improvements and perform notably better than the attention shift score in CAKE and MorphKV (local score only). However, the mean form of the global score performs slightly worse than the other two. The performance of all three methods remains relatively stable when \(\alpha \in [0.8, 0.9]\), achieving notable performance gains. This range is recommended as the optimal hyperparameter setting.}

\subsection{Main Results}
\label{sct:main_res}
In this section, we evaluate the performance of various methods under different budget constraints. We refer to the method that combines the global score (max) with the redundancy score \citep{cai2025r} as \textbf{G-KV}, and we set $\alpha$ to 0.8. \textcolor{black}{Appendix \ref{sct:sim} describes how to combine global score with other methods. The baselines for comparison include StreamingLLM \citep{xiao2023efficient},  MorphKV \citep{ghadiadialogue}, 
and R-KV \citep{cai2025r}. 
We also compared our method with SnapKV \citep{li2024snapkv}. The original SnapKV only supports compression during the prefilling stage; we extended its compression process to the decoding stage.
For StreamingLLM, the budget refers to its window size. For SnapKV and R-KV, the parameters \(s\) and \(w\) are consistent with those used in the previous section, while other parameters follow the settings specified in their respective papers.}
Additionally, we report the average \textbf{Token Retention Ratio}, defined as the ratio of the KV Cache length to the total sequence length. This ratio is calculated exclusively for cases where the model generates the \textbf{correct} answer. \textbf{A lower token retention ratio indicates the model can function properly with longer generation lengths under a fixed budget.}

\begin{figure}[h]
    \centering
    \includegraphics[width=0.97\linewidth]{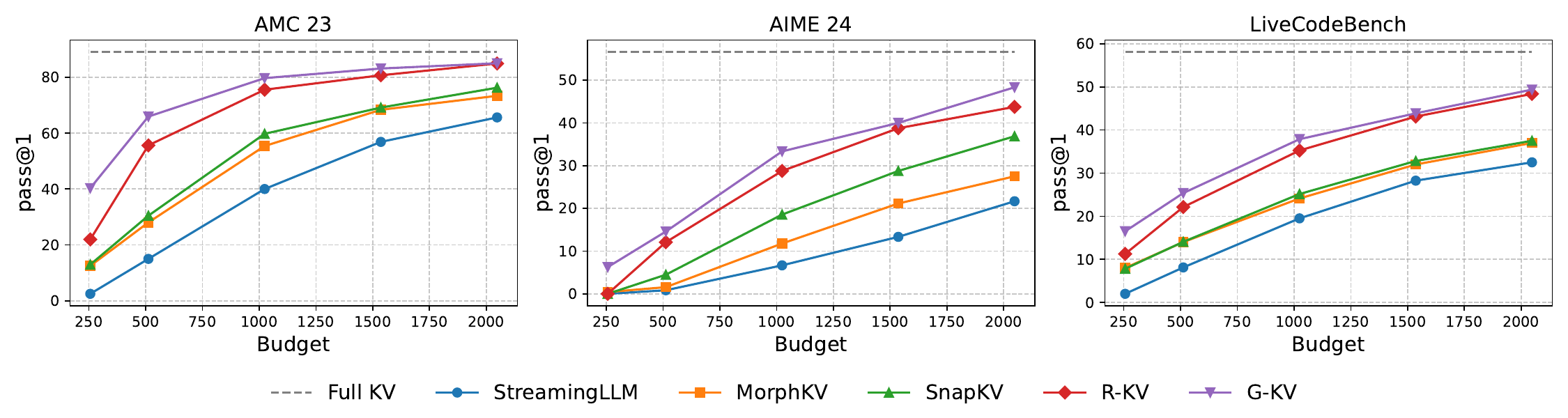}
    \caption{\textcolor{black}{Performance of different compression methods with DeepSeek-R1-Distill Qwen 7B model.}}
    \label{fig:main_pass1}
\end{figure}

\begin{figure}[h]
    \centering
    \includegraphics[width=0.97\linewidth]{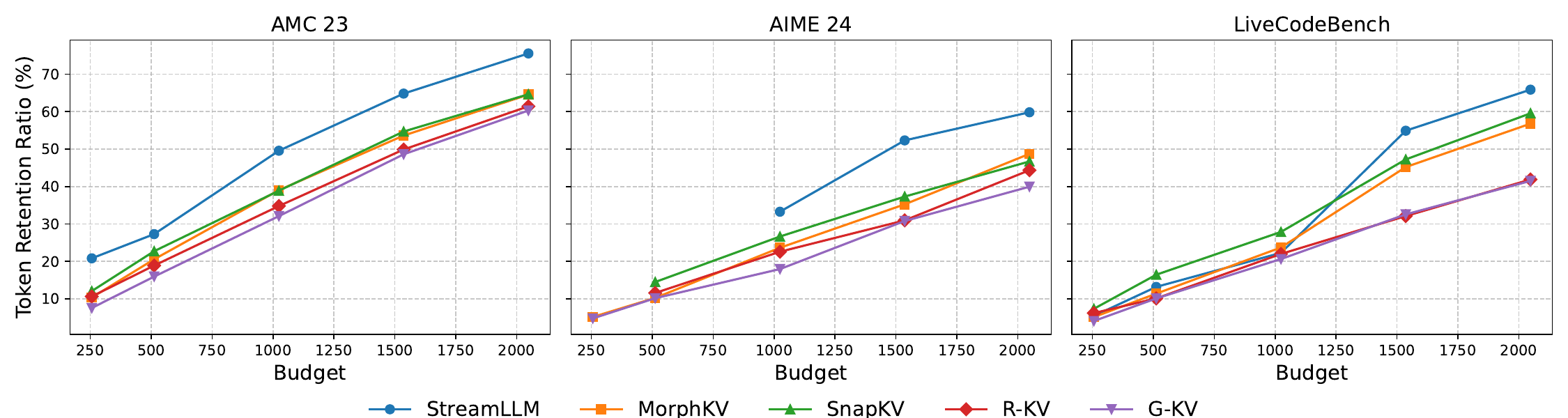}
    \caption{\textcolor{black}{Token retention ratio of different compression methods with DeepSeek-R1-Distill Qwen 7B model.}}
    \label{fig:main_ratio}
\end{figure}

\textbf{Analysis}. \textcolor{black}{As shown in Figure \ref{fig:main_pass1}, our method achieves SOTA performance across most budgets and benchmarks. The fewer the budget tokens, the greater the advantage of our method over others. For the AMC 23 benchmark, our approach achieves nearly a 20\% improvement under a 512-token budget. The results in Figure \ref{fig:main_ratio} also demonstrate that our method achieves the lowest token retention ratio in most scenarios. This indicates that the tokens retained by our method have a higher information density, enabling the model to function effectively on longer sequences.}

\begin{figure}[h]
    \centering
    \includegraphics[width= \linewidth]{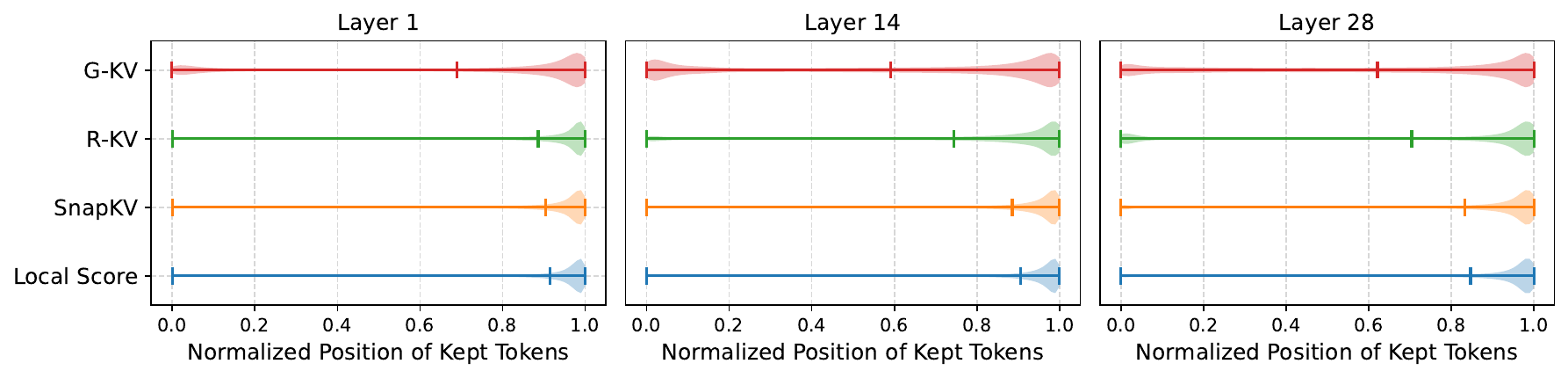}
    \caption{The density distribution of the normalized final retained token positions for different algorithms using DeepSeek-R1-Distill-Qwen-7B. The results are evaluated on the AMC 23 benchmark. The vertical bars in the figure indicate the \textbf{mean values}.}
    \label{fig:qwen_pos_qwen7b}
\end{figure}

Furthermore, we normalize the positions of tokens retained by different algorithms (with budget of 512) within the complete sequence and visualize their density distributions, as shown in Figure \ref{fig:qwen_pos_qwen7b}. The tokens retained by previous methods based on local scores are concentrated towards the end of the sequence. This phenomenon arises because the context near the observation window during each compression step tends to have higher semantic similarity, compounded by the inherent characteristics of RoPE \citep{su2024roformer}. In contrast, when using global scores, the retained token positions are more evenly distributed, allowing more comprehensive information to be preserved. This characteristic may explain why G-KV performs significantly better than other methods in handling longer generation sequences and under lower budget constraints. A cases presented in Appendix \ref{sct:case_study} illustrate this phenomenon more intuitively. \textcolor{black}{We explain this phenomenon as arising from the fact that tokens close to the observation window are more likely to receive higher attention scores. Consequently, methods based on local scores tend to prioritize retaining tokens nearest to the observation window. In contrast, attention to more distant contexts is often intermittent. Global scores allow these intermittently attended but important tokens to be retained, whereas local scores are more prone to evict them when their attention scores temporarily drop.}

\subsection{Results of Training}

In this section, we further present the results obtained using different training methods. We refer to our proposed RL and distillation methods as \textbf{RL-Sparse} and \textbf{Distill}, respectively. For comparison, we include reinforcement learning conducted with generation and training with the Full KV cache, which we refer to as \textbf{RL-Full}.

\textbf{Implementation Details.} The training is based on the DeepSeek-R1-Distill-Qwen-7B model. For RL training, the maximum output length is 4096, with a sampling temperature of 0.6. Each step samples 16 questions, with 8 responses generated per question, yielding 128 trajectories for gradient computation and updates in a single batch (allowing gradient accumulation via micro-batches). RL training runs for 400 steps, rewarding 1 for correct responses and 0 for incorrect ones. For distillation, the maximum sequence length is 4096, with longer sequences truncated. Training is performed for 250 steps (around 1 epoch) with a batch size of 128. The learning rate is set to $1 \times 10^{-6}$. The G-KV method is used for RL-Sparse with a budget of 2048, while other parameters remain as previously mentioned. All evaluations in this section are restricted to an output length of 4096, consistent with the training setup. 

\begin{table}[h]
\centering
\renewcommand{\arraystretch}{1.05}
\begin{tabular}{c c c c c c c c}
\toprule
\textbf{} & \multicolumn{3}{c}{\textbf{AMC 23}} & \multicolumn{3}{c}{\textbf{AIME 24}} \\
\cmidrule(lr){2-4} \cmidrule(lr){5-7}
\textbf{Budget} & \textbf{512} & \textbf{1024} & \textbf{2048} & \textbf{512} & \textbf{1024} & \textbf{2048} \\
\midrule
\multirow{1}{*}{\textbf{Untrained}}  
    & 45.00 & 54.21 & 59.84 
    & 11.56 & 18.64 & 23.02 \\
\midrule
\multirow{2}{*}{\textbf{Distill}}    
    & 47.89 & 56.48 & 61.48 
    & \textbf{14.27} & 21.56 & 24.79 \\
    & \textcolor{green!60!black}{(+2.89)} & \textcolor{green!60!black}{(+2.27)} & \textcolor{green!60!black}{(+1.64)} 
    & \textcolor{green!60!black}{(+2.71)} & \textcolor{green!60!black}{(+2.92)} & \textcolor{green!60!black}{(+1.77)} \\
\multirow{2}{*}{\textbf{RL-Full}}    
    & 47.65 & 56.79 & 63.82 
    & 12.18 & 21.14 & 25.93 \\
    & \textcolor{green!60!black}{(+2.65)} & \textcolor{green!60!black}{(+2.58)} & \textcolor{green!60!black}{(+3.98)} 
    & \textcolor{green!60!black}{(+0.62)} & \textcolor{green!60!black}{(+2.50)} & \textcolor{green!60!black}{(+2.91)} \\
\multirow{2}{*}{\textbf{RL-Sparse}}  
    & \textbf{51.01} & \textbf{61.71} & \textbf{67.65} 
    & 13.75 & \textbf{22.18} & \textbf{26.66} \\
    & \textcolor{green!60!black}{(+6.01)} & \textcolor{green!60!black}{(+7.50)} & \textcolor{green!60!black}{(+7.81)} 
    & \textcolor{green!60!black}{(+2.19)} & \textcolor{green!60!black}{(+3.54)} & \textcolor{green!60!black}{(+3.64)} \\
\bottomrule
\end{tabular}
\caption{Pass@1 comparison across different training methods and budgets on AMC 23 and AIME 24.}
\label{tab:train_results}
\end{table}

\textbf{Analysis}. We evaluated the trained models under different budgets, with the results summarized in Table \ref{tab:train_results}. \textbf{RL-Sparse} achieves the best performance across most settings, significantly outperforming models trained with the Full KV cache. By directly optimizing the policy $\pi'_\theta$, \textbf{RL-Sparse} minimizes the training-inference discrepancy, resulting in superior performance. In contrast, \textbf{RL-Full} shows moderate gains but is hindered by the mismatch between its training policy $\pi_\theta$ and the inference policy. The distillation method effectively enables $\pi'_\theta$ under constrained KV cache to approximate $\pi_\theta$, offering a practical alternative for scenarios where verifiable reward functions are difficult to design. Additional training information and analysis are provided in Appendix \ref{sct:train_info}.

\subsection{Efficiency Analysis}
\label{sct:eff_analysis}

In this section, we analyze the efficiency of the KV cache compression algorithm. Throughput is used as the evaluation metric, calculated as the total number of valid tokens generated (excluding padding tokens) divided by the time consumed. We extracted 1,024 mathematical problems from the DeepScaleR-40k dataset and conducted inference with varying batch sizes, using a maximum output length of 16k. All experiments were performed on a single A100 GPU.

\begin{table}[h]
\centering
\renewcommand{\arraystretch}{1.05}
\setlength{\tabcolsep}{10pt}
\begin{tabular}{c c c c c c c}
\toprule
\textbf{} & \multicolumn{3}{c}{\textbf{DeepSeek-Qwen-Distill-7B}} & \multicolumn{3}{c}{\textbf{DeepSeek-Llama-Distill-8B}} \\
\cmidrule(lr){2-4} \cmidrule(lr){5-7}
\textbf{Batch Size} & \textbf{32} & \textbf{64} & \textbf{128} & \textbf{16} & \textbf{32} & \textbf{64} \\
\midrule
\textbf{Full-KV} 
    & 62.41 & OOM & OOM 
    & 31.08 & OOM & OOM \\
\midrule
\textbf{R-KV (Budget 2048)} 
    & 172.44 & 203.66 & 238.43 
    & 82.33 & 98.01 & 111.89 \\
\midrule
\textbf{G-KV (Budget 512)} 
    & 261.32 & 475.59 & 760.74 
    & 158.43 & 517.30 & 612.60 \\
\textbf{G-KV (Budget 1024)} 
    & 212.93 & 367.96 & 448.35 
    & 118.10 & 193.56 & 258.18 \\
\textbf{G-KV (Budget 2048)} 
    & 170.64 & 221.23 & 248.22 
    & 93.91 & 118.82 & 154.52 \\
\bottomrule
\end{tabular}
\caption{Throughput comparison (tokens/s). OOM refers to the occurrence of an Out of Memory error, indicating insufficient GPU memory. }
\label{tab:throughput_comparison}
\end{table}

\textbf{Analysis.} As shown in Table \ref{tab:throughput_comparison}, our method achieves a significant improvement in throughput compared to Full-KV under the same batch size. For DeepSeek-Qwen-Distill-7B, throughput improves by 4.18$\times$, 3.41$\times$, and 2.73$\times$ under KV cache budgets of 512, 1024, and 2048, respectively. Similarly, DeepSeek-Qwen-Llama-8B achieves throughput gains of 5.09$\times$, 3.79$\times$, and 3.02$\times$ under the same budgets. Naturally, the reduced memory requirements of the KV cache allow inference with larger batch sizes. For these two models, the throughput of GKV reaches up to 12.18$\times$ and 19.7$\times$ that of Full-KV, respectively. 
\begin{wrapfigure}{r}{0.45\linewidth}
\centering
\includegraphics[width=0.9\linewidth]{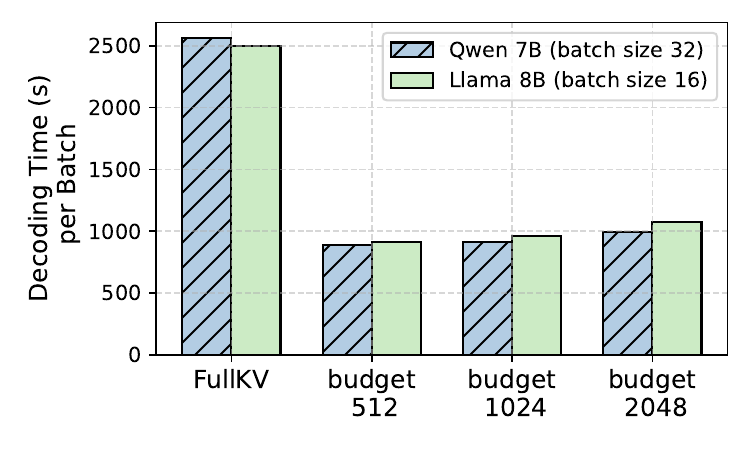}
\caption{Decoding time comparison.}
\label{fig:time_v_budget}
\end{wrapfigure}
Additionally, we conducted experiments on the throughput of R-KV. Since our method operates within the same framework as R-KV and introduces minimal additional computation, the difference in throughput between the two methods is negligible.

In addition to throughput, decoding time is a critical factor influencing user experience in practical applications. As shown in Figure \ref{fig:time_v_budget}, the decoding time under different KV cache compression budgets is similar and, at the same batch size, is approximately 40\% of that of Full-KV. Further comparisons and analyses of decoding time are provided in Appendix \ref{sct:time_cmp}. Additionally, we conducted an analysis of memory efficiency. Under the 16k context setting, our method achieves approximately a 90\% reduction in KV cache memory usage. Detailed results are provided in Appendix \ref{sct:memory_analysis}.

\begin{minipage}[t]{0.48\textwidth}
\centering
\begin{tabular}{lccc}
\toprule
\textbf{Budget} & \textbf{512} & \textbf{1024} & \textbf{2048} \\
\midrule
Local score  & 46.1  & 74.5  & 117.6 \\
Global score & 50.5  & 79.4  & 121.4 \\
\bottomrule
\end{tabular}
\captionof{table}{\textcolor{black}{Average Compression Time (ms)}}
\label{tab:compress_time}
\end{minipage}%
\hfill
\begin{minipage}[t]{0.48\textwidth}
\centering
\begin{tabular}{lccc}
\toprule
\textbf{Budget} & \textbf{512} & \textbf{1024} & \textbf{2048} \\
\midrule
Local score  & 0.77\% & 1.10\% & 1.59\% \\
Global score & 0.83\% & 1.20\% & 1.57\% \\
\bottomrule
\end{tabular}
\captionof{table}{\textcolor{black}{Compression Time Ratio (\%)}}
\label{tab:compress_ratio}
\end{minipage}

\textcolor{black}{We use DeepSeek-Distill Qwen-7B to measure the average compression time per step for global score and local score under a batch size of 32. The experimental results are shown in Table \ref{tab:compress_time}. Global score introduces an additional delay of approximately 5 ms per compression step. However, this delay is negligible in the context of the entire decoding process. Table \ref{tab:compress_ratio} presents the proportion of compression time relative to the total decoding time, showing that the compression process for both global score and local score accounts for only about 1\% of the total time.}

\section{Conclusion}

In this paper, we propose G-KV, a KV cache compression method that integrates local and historical attention scores to assess token importance globally. In addition, post-training techniques, including reinforcement learning and distillation, are introduced to adapt LLMs to compressed KV cache settings. Experiments on AMC-23 and AIME-24 benchmarks confirm effectiveness of G-KV. G-KV significantly reduces memory and computational bottlenecks, enabling efficient and scalable reasoning for LLMs.




\bibliography{ref}

@article{zhang2023h2o,
  title={H2o: Heavy-hitter oracle for efficient generative inference of large language models},
  author={Zhang, Zhenyu and Sheng, Ying and Zhou, Tianyi and Chen, Tianlong and Zheng, Lianmin and Cai, Ruisi and Song, Zhao and Tian, Yuandong and R{\'e}, Christopher and Barrett, Clark and others},
  journal={Advances in Neural Information Processing Systems},
  volume={36},
  pages={34661--34710},
  year={2023}
}

@inproceedings{ghadiadialogue,
  title={Dialogue Without Limits: Constant-Sized KV Caches for Extended Response in LLMs},
  author={Ghadia, Ravi and Kumar, Avinash and Jain, Gaurav and Nair, Prashant J and Das, Poulami},
  booktitle={Forty-second International Conference on Machine Learning},
year={2025}
}

@article{kim2025kvzip,
  title={KVzip: Query-Agnostic KV Cache Compression with Context Reconstruction},
  author={Kim, Jang-Hyun and Kim, Jinuk and Kwon, Sangwoo and Lee, Jae W and Yun, Sangdoo and Song, Hyun Oh},
  journal={arXiv preprint arXiv:2505.23416},
  year={2025}
}

@article{li2024snapkv,
  title={Snapkv: Llm knows what you are looking for before generation},
  author={Li, Yuhong and Huang, Yingbing and Yang, Bowen and Venkitesh, Bharat and Locatelli, Acyr and Ye, Hanchen and Cai, Tianle and Lewis, Patrick and Chen, Deming},
  journal={Advances in Neural Information Processing Systems},
  volume={37},
  pages={22947--22970},
  year={2024}
}

@article{cai2025r,
  title={R-KV: Redundancy-aware KV Cache Compression for Training-Free Reasoning Models Acceleration},
  author={Cai, Zefan and Xiao, Wen and Sun, Hanshi and Luo, Cheng and Zhang, Yikai and Wan, Ke and Li, Yucheng and Zhou, Yeyang and Chang, Li-Wen and Gu, Jiuxiang and others},
  journal={Advances in Neural Information Processing Systems},
  year={2025}
}

@article{xiao2023efficient,
  title={Efficient streaming language models with attention sinks},
  author={Xiao, Guangxuan and Tian, Yuandong and Chen, Beidi and Han, Song and Lewis, Mike},
  journal={arXiv preprint arXiv:2309.17453},
  year={2023}
}

@article{team2025kimi,
  title={Kimi k1. 5: Scaling reinforcement learning with llms},
  author={Team, Kimi and Du, Angang and Gao, Bofei and Xing, Bowei and Jiang, Changjiu and Chen, Cheng and Li, Cheng and Xiao, Chenjun and Du, Chenzhuang and Liao, Chonghua and others},
  journal={arXiv preprint arXiv:2501.12599},
  year={2025}
}

@article{yang2025qwen3,
  title={Qwen3 technical report},
  author={Yang, An and Li, Anfeng and Yang, Baosong and Zhang, Beichen and Hui, Binyuan and Zheng, Bo and Yu, Bowen and Gao, Chang and Huang, Chengen and Lv, Chenxu and others},
  journal={arXiv preprint arXiv:2505.09388},
  year={2025}
}

@misc{deepscaler2025,
  title={DeepScaleR: Surpassing O1-Preview with a 1.5B Model by Scaling RL},
  author={Michael Luo and Sijun Tan and Justin Wong and Xiaoxiang Shi and William Y. Tang and Manan Roongta and Colin Cai and Jeffrey Luo and Tianjun Zhang and Li Erran Li and Raluca Ada Popa and Ion Stoica},
  year={2025},
  howpublished={https://pretty-radio-b75.notion.site/DeepScaleR-Surpassing-O1-Preview-with-a-1-5B-Model-by-Scaling-RL-19681902c1468005bed8ca303013a4e2},
  note={Notion Blog}
}

@article{chen2021evaluating,
  title={Evaluating large language models trained on code},
  author={Chen, Mark and Tworek, Jerry and Jun, Heewoo and Yuan, Qiming and Pinto, Henrique Ponde De Oliveira and Kaplan, Jared and Edwards, Harri and Burda, Yuri and Joseph, Nicholas and Brockman, Greg and others},
  journal={arXiv preprint arXiv:2107.03374},
  year={2021}
}

@article{grattafiori2024llama,
  title={The llama 3 herd of models},
  author={Grattafiori, Aaron and Dubey, Abhimanyu and Jauhri, Abhinav and Pandey, Abhinav and Kadian, Abhishek and Al-Dahle, Ahmad and Letman, Aiesha and Mathur, Akhil and Schelten, Alan and Vaughan, Alex and others},
  journal={arXiv preprint arXiv:2407.21783},
  year={2024}
}

@misc{aops2023amc8,
  author       = {{AoPS}},
  title        = {2023 AMC 8 -- AoPS Wiki},
  year         = {2023},
  url          = {https://artofproblemsolving.com/wiki/index.php/2023_AMC_8},
  note         = {Accessed: September 18, 2025}
}

@misc{aops2024aimei,
  author       = {{AoPS}},
  title        = {2024 AIME I -- AoPS Wiki},
  year         = {2024},
  url          = {https://artofproblemsolving.com/wiki/index.php/2024_AIME_I},
  note         = {Accessed: September 18, 2025}
}

@article{team2024qwen2,
  title={Qwen2 technical report},
  author={Team, Qwen},
  journal={arXiv preprint arXiv:2407.10671},
  year={2024}
}

@article{guo2025deepseek,
  title={Deepseek-r1: Incentivizing reasoning capability in llms via reinforcement learning},
  author={Guo, Daya and Yang, Dejian and Zhang, Haowei and Song, Junxiao and Zhang, Ruoyu and Xu, Runxin and Zhu, Qihao and Ma, Shirong and Wang, Peiyi and Bi, Xiao and others},
  journal={arXiv preprint arXiv:2501.12948},
  year={2025}
}

@article{cai2024pyramidkv,
  title={Pyramidkv: Dynamic kv cache compression based on pyramidal information funneling},
  author={Cai, Zefan and Zhang, Yichi and Gao, Bofei and Liu, Yuliang and Li, Yucheng and Liu, Tianyu and Lu, Keming and Xiong, Wayne and Dong, Yue and Hu, Junjie and others},
  journal={arXiv preprint arXiv:2406.02069},
  year={2024}
}

@article{kim2023compressed,
  title={Compressed context memory for online language model interaction},
  author={Kim, Jang-Hyun and Yeom, Junyoung and Yun, Sangdoo and Song, Hyun Oh},
  journal={arXiv preprint arXiv:2312.03414},
  year={2023}
}

@article{shah2025rethinking,
  title={Rethinking reflection in pre-training},
  author={Shah, Darsh J and Rushton, Peter and Singla, Somanshu and Parmar, Mohit and Smith, Kurt and Vanjani, Yash and Vaswani, Ashish and Chaluvaraju, Adarsh and Hojel, Andrew and Ma, Andrew and others},
  journal={arXiv preprint arXiv:2504.04022},
  year={2025}
}

@article{muennighoff2025s1,
  title={s1: Simple test-time scaling},
  author={Muennighoff, Niklas and Yang, Zitong and Shi, Weijia and Li, Xiang Lisa and Fei-Fei, Li and Hajishirzi, Hannaneh and Zettlemoyer, Luke and Liang, Percy and Cand{\`e}s, Emmanuel and Hashimoto, Tatsunori},
  journal={arXiv preprint arXiv:2501.19393},
  year={2025}
}

@article{jain2024livecodebench,
  title={Livecodebench: Holistic and contamination free evaluation of large language models for code},
  author={Jain, Naman and Han, King and Gu, Alex and Li, Wen-Ding and Yan, Fanjia and Zhang, Tianjun and Wang, Sida and Solar-Lezama, Armando and Sen, Koushik and Stoica, Ion},
  journal={arXiv preprint arXiv:2403.07974},
  year={2024}
}

@inproceedings{funot,
  title={Not All Heads Matter: A Head-Level KV Cache Compression Method with Integrated Retrieval and Reasoning},
  author={Fu, Yu and Cai, Zefan and Asi, Abedelkadir and Xiong, Wayne and Dong, Yue and Xiao, Wen},
  booktitle={The Thirteenth International Conference on Learning Representations},
  year={2024}
}

@article{dong2024get,
  title={Get more with less: Synthesizing recurrence with kv cache compression for efficient llm inference},
  author={Dong, Harry and Yang, Xinyu and Zhang, Zhenyu and Wang, Zhangyang and Chi, Yuejie and Chen, Beidi},
  journal={arXiv preprint arXiv:2402.09398},
  year={2024}
}

@article{chang2024palu,
  title={Palu: Compressing kv-cache with low-rank projection},
  author={Chang, Chi-Chih and Lin, Wei-Cheng and Lin, Chien-Yu and Chen, Chong-Yan and Hu, Yu-Fang and Wang, Pei-Shuo and Huang, Ning-Chi and Ceze, Luis and Abdelfattah, Mohamed S and Wu, Kai-Chiang},
  journal={arXiv preprint arXiv:2407.21118},
  year={2024}
}

@inproceedings{liu2024cachegen,
  title={Cachegen: Kv cache compression and streaming for fast large language model serving},
  author={Liu, Yuhan and Li, Hanchen and Cheng, Yihua and Ray, Siddhant and Huang, Yuyang and Zhang, Qizheng and Du, Kuntai and Yao, Jiayi and Lu, Shan and Ananthanarayanan, Ganesh and others},
  booktitle={Proceedings of the ACM SIGCOMM 2024 Conference},
  pages={38--56},
  year={2024}
}

@inproceedings{sheng2023flexgen,
  title={Flexgen: High-throughput generative inference of large language models with a single gpu},
  author={Sheng, Ying and Zheng, Lianmin and Yuan, Binhang and Li, Zhuohan and Ryabinin, Max and Chen, Beidi and Liang, Percy and R{\'e}, Christopher and Stoica, Ion and Zhang, Ce},
  booktitle={International Conference on Machine Learning},
  pages={31094--31116},
  year={2023},
  organization={PMLR}
}

@article{yao2022zeroquant,
  title={Zeroquant: Efficient and affordable post-training quantization for large-scale transformers},
  author={Yao, Zhewei and Yazdani Aminabadi, Reza and Zhang, Minjia and Wu, Xiaoxia and Li, Conglong and He, Yuxiong},
  journal={Advances in neural information processing systems},
  volume={35},
  pages={27168--27183},
  year={2022}
}

@article{liu2024minicache,
  title={Minicache: Kv cache compression in depth dimension for large language models},
  author={Liu, Akide and Liu, Jing and Pan, Zizheng and He, Yefei and Haffari, Gholamreza and Zhuang, Bohan},
  journal={Advances in Neural Information Processing Systems},
  volume={37},
  pages={139997--140031},
  year={2024}
}

@inproceedings{nawrot2024dynamic,
  title={Dynamic Memory Compression: Retrofitting LLMs for Accelerated Inference},
  author={Nawrot, Piotr and {\L}a{\'n}cucki, Adrian and Chochowski, Marcin and Tarjan, David and Ponti, Edoardo},
  booktitle={International Conference on Machine Learning},
  pages={37396--37412},
  year={2024},
  organization={PMLR}
}

@article{cui2025entropy,
  title={The entropy mechanism of reinforcement learning for reasoning language models},
  author={Cui, Ganqu and Zhang, Yuchen and Chen, Jiacheng and Yuan, Lifan and Wang, Zhi and Zuo, Yuxin and Li, Haozhan and Fan, Yuchen and Chen, Huayu and Chen, Weize and others},
  journal={arXiv preprint arXiv:2505.22617},
  year={2025}
}

@article{wang2025beyond,
  title={Beyond the 80/20 rule: High-entropy minority tokens drive effective reinforcement learning for llm reasoning},
  author={Wang, Shenzhi and Yu, Le and Gao, Chang and Zheng, Chujie and Liu, Shixuan and Lu, Rui and Dang, Kai and Chen, Xionghui and Yang, Jianxin and Zhang, Zhenru and others},
  journal={arXiv preprint arXiv:2506.01939},
  year={2025}
}

@article{liao2025enhancing,
  title={Enhancing Efficiency and Exploration in Reinforcement Learning for LLMs},
  author={Liao, Mengqi and Xi, Xiangyu and Chen, Ruinian and Leng, Jia and Hu, Yangen and Zeng, Ke and Liu, Shuai and Wan, Huaiyu},
  journal={arXiv preprint arXiv:2505.18573},
  year={2025}
}

@article{shannon1948mathematical,
  title={A mathematical theory of communication},
  author={Shannon, Claude E},
  journal={The Bell system technical journal},
  volume={27},
  number={3},
  pages={379--423},
  year={1948},
  publisher={Nokia Bell Labs}
}

@article{su2024roformer,
  title={Roformer: Enhanced transformer with rotary position embedding},
  author={Su, Jianlin and Ahmed, Murtadha and Lu, Yu and Pan, Shengfeng and Bo, Wen and Liu, Yunfeng},
  journal={Neurocomputing},
  volume={568},
  pages={127063},
  year={2024},
  publisher={Elsevier}
}

@inproceedings{yang2024pyramidinfer,
  title={PyramidInfer: Pyramid KV Cache Compression for High-throughput LLM Inference},
  author={Yang, Dongjie and Han, Xiaodong and Gao, Yan and Hu, Yao and Zhang, Shilin and Zhao, Hai},
  booktitle={Findings of the Association for Computational Linguistics ACL 2024},
  pages={3258--3270},
  year={2024}
}

@inproceedings{qin2025cake,
  title={CAKE: Cascading and Adaptive KV Cache Eviction with Layer Preferences},
  author={Qin, Ziran and Cao, Yuchen and Lin, Mingbao and Hu, Wen and Fan, Shixuan and Cheng, Ke and Lin, Weiyao and Li, Jianguo},
  booktitle={The Thirteenth International Conference on Learning Representations},
  year={2025}
}

@article{song2025reasoning,
  title={Reasoning Path Compression: Compressing Generation Trajectories for Efficient LLM Reasoning},
  author={Song, Jiwon and Jo, Dongwon and Kim, Yulhwa and Kim, Jae-Joon},
  journal={arXiv preprint arXiv:2505.13866},
  year={2025}
}

@article{feng2024ada,
  title={Ada-kv: Optimizing kv cache eviction by adaptive budget allocation for efficient llm inference},
  author={Feng, Yuan and Lv, Junlin and Cao, Yukun and Xie, Xike and Zhou, S Kevin},
  journal={arXiv preprint arXiv:2407.11550},
  year={2024}
}

@article{kullback1951information,
  title={On information and sufficiency},
  author={Kullback, Solomon and Leibler, Richard A},
  journal={The annals of mathematical statistics},
  volume={22},
  number={1},
  pages={79--86},
  year={1951},
  publisher={JSTOR}
}

@article{hinton2015distilling,
  title={Distilling the knowledge in a neural network},
  author={Hinton, Geoffrey and Vinyals, Oriol and Dean, Jeff},
  journal={arXiv preprint arXiv:1503.02531},
  year={2015}
}

@article{vaswani2017attention,
  title={Attention is all you need},
  author={Vaswani, Ashish and Shazeer, Noam and Parmar, Niki and Uszkoreit, Jakob and Jones, Llion and Gomez, Aidan N and Kaiser, {\L}ukasz and Polosukhin, Illia},
  journal={Advances in neural information processing systems},
  volume={30},
  year={2017}
}

@article{ainslie2023gqa,
  title={Gqa: Training generalized multi-query transformer models from multi-head checkpoints},
  author={Ainslie, Joshua and Lee-Thorp, James and De Jong, Michiel and Zemlyanskiy, Yury and Lebr{\'o}n, Federico and Sanghai, Sumit},
  journal={arXiv preprint arXiv:2305.13245},
  year={2023}
}

@article{shazeer2019fast,
  title={Fast transformer decoding: One write-head is all you need},
  author={Shazeer, Noam},
  journal={arXiv preprint arXiv:1911.02150},
  year={2019}
}

@article{shao2024deepseekmath,
  title={Deepseekmath: Pushing the limits of mathematical reasoning in open language models},
  author={Shao, Zhihong and Wang, Peiyi and Zhu, Qihao and Xu, Runxin and Song, Junxiao and Bi, Xiao and Zhang, Haowei and Zhang, Mingchuan and Li, YK and Wu, Yang and others},
  journal={arXiv preprint arXiv:2402.03300},
  year={2024}
}

@article{yu2025dapo,
  title={Dapo: An open-source llm reinforcement learning system at scale},
  author={Yu, Qiying and Zhang, Zheng and Zhu, Ruofei and Yuan, Yufeng and Zuo, Xiaochen and Yue, Yu and Dai, Weinan and Fan, Tiantian and Liu, Gaohong and Liu, Lingjun and others},
  journal={arXiv preprint arXiv:2503.14476},
  year={2025}
}

@inproceedings{chensepllm,
  title={SepLLM: Accelerate Large Language Models by Compressing One Segment into One Separator},
  author={Chen, Guoxuan and Shi, Han and Li, Jiawei and Gao, Yihang and Ren, Xiaozhe and Chen, Yimeng and Jiang, Xin and Li, Zhenguo and Liu, Weiyang and Huang, Chao},
  booktitle={Forty-second International Conference on Machine Learning},
  year={2025}
}

@article{li2024survey,
  title={A survey on large language model acceleration based on kv cache management},
  author={Li, Haoyang and Li, Yiming and Tian, Anxin and Tang, Tianhao and Xu, Zhanchao and Chen, Xuejia and Hu, Nicole and Dong, Wei and Li, Qing and Chen, Lei},
  journal={arXiv preprint arXiv:2412.19442},
  year={2024}
}

@article{wei2022chain,
  title={Chain-of-thought prompting elicits reasoning in large language models},
  author={Wei, Jason and Wang, Xuezhi and Schuurmans, Dale and Bosma, Maarten and Xia, Fei and Chi, Ed and Le, Quoc V and Zhou, Denny and others},
  journal={Advances in neural information processing systems},
  volume={35},
  pages={24824--24837},
  year={2022}
}
\bibliographystyle{iclr2026_conference}

\appendix

\section{Use of LLMs}
We utilized ChatGPT-4o \footnote{https://chatgpt.com
} to refine the content based on our original writing. All revised text was subsequently reviewed and verified by us. The architecture of the code was designed by our team, with Claude-4 \footnote{https://claude.ai
} assisting in the implementation of certain functional components. All code has undergone comprehensive testing to ensure its reliability.

\section{The Sparse Nature of Attention}

\label{sct:sparse_nature}
In this section, we analyze the sparsity of attention scores. Specifically, let the maximum attention score in a sequence be $s_{\max}$. We define $p \times s_{\max}$, where $p \in (0,1)$, as a threshold. Tokens with attention scores below this threshold are considered to receive minimal attention. The \textit{sparsity}  is defined as the proportion of tokens with attention scores below the threshold relative to the total number of tokens in the sequence.

For the full KV cache, we compute the attention scores between the query states of the last 16 tokens and the key states of all preceding tokens. However, we only evaluate the sparsity of the last 512 tokens. For KV cache compression algorithms, we calculate the attention scores between the query states of tokens in the last observation window and the key states retained in the kept KV cache. For all KV cache compression algorithms, we set the budget to 512, the observation window size to 16, and the compression interval to 128.

\begin{figure}[h]
    \centering
    \begin{subfigure}{0.95\linewidth}
        \centering
        \includegraphics[width=\linewidth]{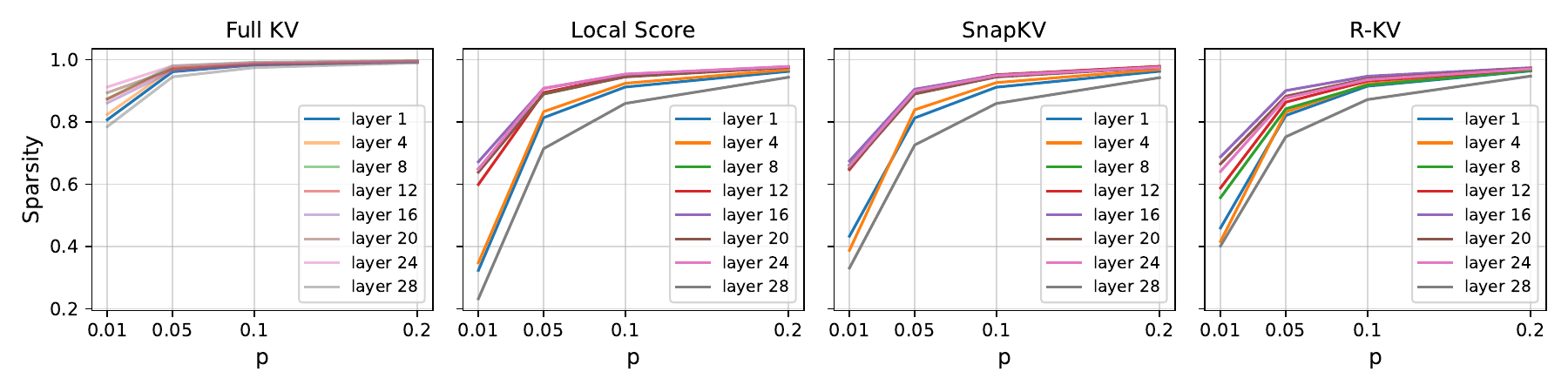}
        \caption{Sparsity of attention in DeepSeek-R1-Distill-Qwen-7B.}
        \label{fig:qwen_sparsity}
    \end{subfigure}

    \vspace{1em} 

    \begin{subfigure}{0.95\linewidth}
        \centering
        \includegraphics[width=\linewidth]{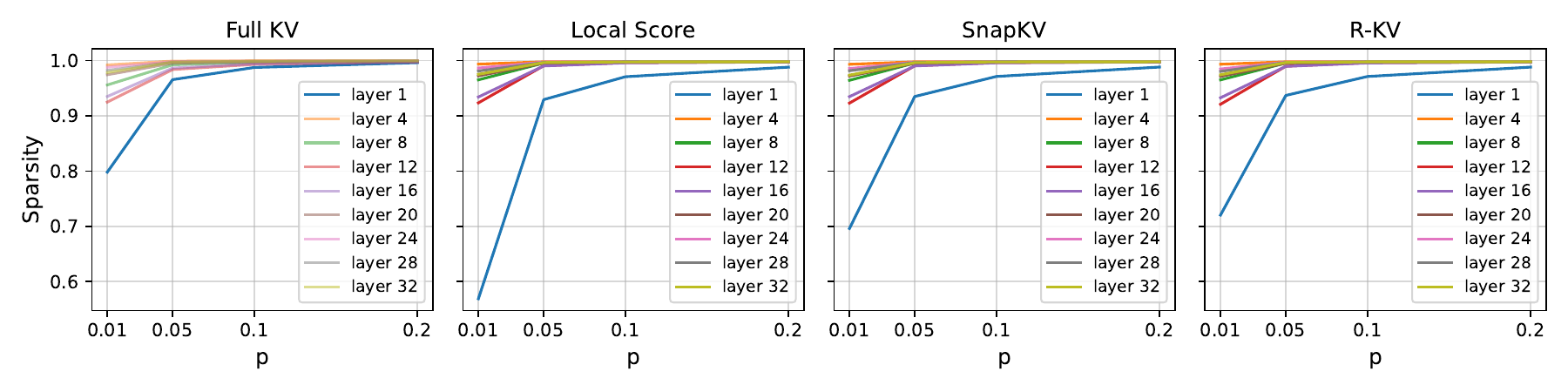}
        \caption{Sparsity of attention in DeepSeek-R1-Distill-LLaMA-8B.}
        \label{fig:llama_sparsity}
    \end{subfigure}

    \caption{Comparison of attention sparsity. The horizontal axis represents the coefficient $p$ multiplied by the maximum attention score to calculate the threshold. The vertical axis represents the sparsity.}
    \label{fig:attn_sparse_level}
\end{figure}

Figure \ref{fig:attn_sparse_level} illustrates the sparsity levels across different models and layers. For full KV cache, the attention scores of most layers exhibit extremely high sparsity. In the majority of layers, over 90\% of tokens have attention scores lower than 1\% of the maximum score. This observation indicates that most tokens are not attended to by the last few tokens, which also serves as the primary motivation behind the design of most existing KV cache compression algorithms \citep{zhang2023h2o,cai2025r}. 

Furthermore, we conducted an analysis of sparsity when applying KV cache compression algorithms. Although the sparsity decreases significantly compared to the full KV cache, notable sparsity still persists. For DeepSeek-R1-Distill-Qwen-7B, many layers still exhibit over 80\% of tokens having attention scores below 5\% of the maximum score. Similarly, for DeepSeek-R1-Distill-LLaMa-8B, with the exception of the first layer, more than 90\% of tokens in other layers have attention scores below 1\% of the maximum score. \textbf{This indicates that even after KV cache compression, the attention scores between the compressed KV cache and the observation window still maintain a high degree of sparsity. This means that each observation window still only attends to a subset of tokens within the compressed KV cache.}

\section{Distillation-Like Training with Sparse Attention Mask}

\label{sct:distill}
As discussed previously, when defining the reward for outputs becomes challenging, reinforcement learning (RL) may no longer be applicable. In such scenarios, alternative training methods need to be explored. In this section, we propose a distillation-based approach.

Specifically, we sample outputs $y \sim \pi_\theta(y|x)$ from $\pi_\theta$. Then, we simulate the execution of our KV cache eviction algorithm to determine which tokens would be evicted during the generation process, thereby constructing a corresponding sparse attention mask. We train $\pi'_\phi$ ($\phi$ initialized as $\theta$) with the sparse attention mask  through a soft target loss \citep{hinton2015distilling}:

\begin{equation}
\mathcal{L}(\phi) = \tau^2\,\text{KL}\!\left(\pi_\theta(y|x)\,\big\|\,\pi'_\phi(y|x)\right),
\end{equation}

where $\tau$ represents the sampling temperature, and $\text{KL}(\cdot\|\cdot)$ denotes the Kullback-Leibler (KL) divergence \citep{kullback1951information}. Minimizing this objective allows the distribution of the policy employing KV cache compression to approximate that of the full KV cache policy. This training approach also enables the model to adapt to the sparse attention.

\section{Integrating the Global Score with Other Methods}
\label{sct:sim}

\begin{figure}[h]
    \centering
    \includegraphics[width=0.95\linewidth, trim=0 140 20 140, clip]{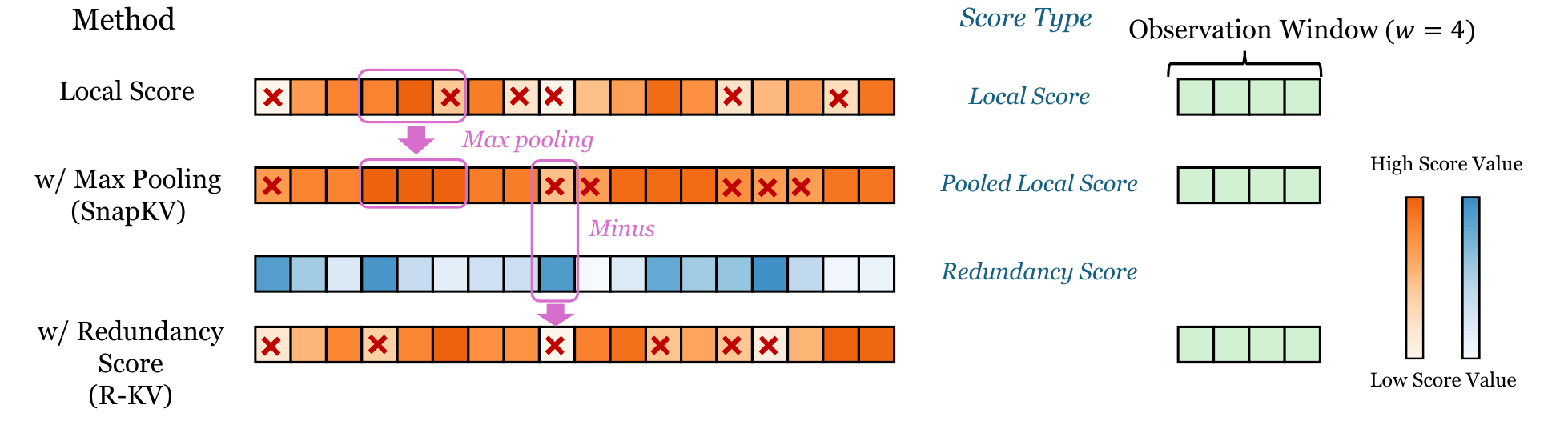}
    \caption{This figure illustrates the computation process of local score, SnapKV and R-KV. Each block represents the KV cache of a token, with the block's color indicating its score (darker color represent higher scores). }
    \label{fig:rkv}
\end{figure}

\textcolor{black}{The global score is an inherently versatile technique that can be seamlessly integrated into other methods. We take SnapKV \citep{li2024snapkv} and R-KV \citep{cai2025r} as examples to demonstrate the results after incorporating the global score.} The schematic diagrams of these methods are illustrated in Figure \ref{fig:rkv}.

\begin{figure}
    \centering
    \includegraphics[width=0.8\linewidth]{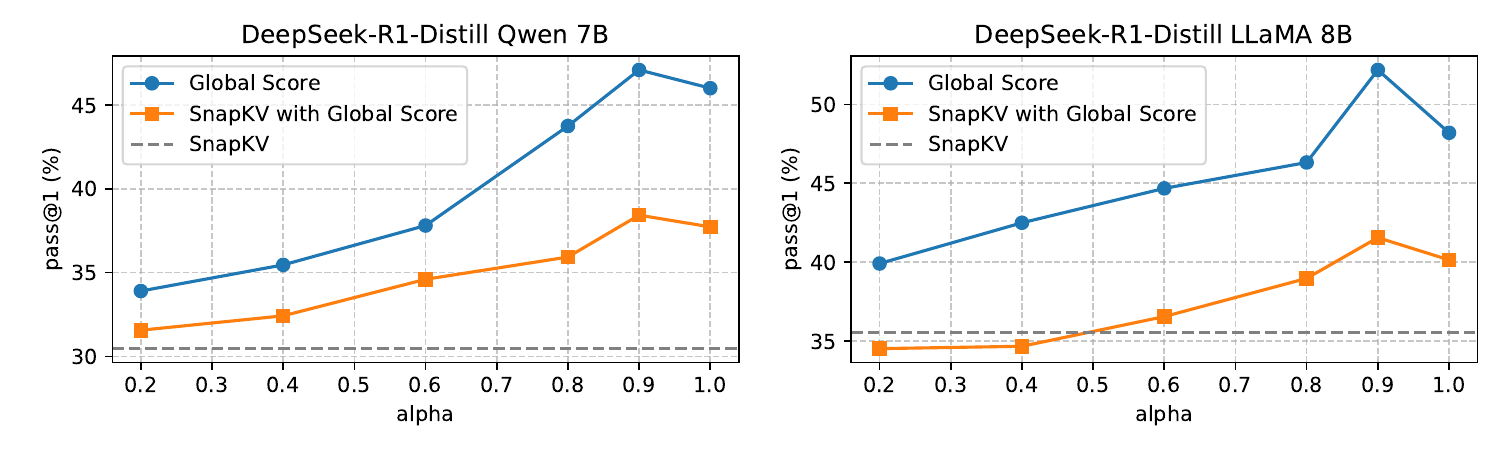}
    \caption{\textcolor{black}{The performance on AMC 23 of SnapKV with Global Score.}}
    \label{fig:snapkv}
\end{figure}

SnapKV introduces sequence-wise max-pooling helps to retain more detailed information from the prompt. When the global score is integrated with SnapKV, it suffices to replace the local score utilized by SnapKV with the global score. \textcolor{black}{ Figure \ref{fig:snapkv} illustrates the effect of combining SnapKV with the global score (max). Replacing the local score with the global score effectively improves the performance of SnapKV; however, it does not surpass the performance of using the global score alone. This may be attributed to the pooling mechanism of SnapKV, which is designed for prefilling-stage compression to retain more detailed information from the prompt. However, during the decoding phase, this design may hinder the eviction of less important tokens.}

R-KV proposed a redundancy score to identify redundant tokens in the KV cache. By removing these redundant tokens, it becomes possible to retain more informative content within a limited KV cache budget. Specifically, the cosine similarity between the Key states, $\mathbf {K}\in \mathbb R^{h_\text{kv} \times l \times d}$, is calculated as follows: 

\[
 \overline{\mathbf K}_{i,j}=\frac{\mathbf K_{i,j}}{\|\mathbf{K}_{i,j}\|_2 +\epsilon},
\]

\[
\mathbf C_i =  \overline {\mathbf K}_i(\overline{\mathbf K}_i^T).
\]

Here, $\mathbf{C} \in \mathbb{R}^{h_\text{kv} \times l \times l}$, \(\mathbf{C}_i\) represents the cosine similarity between the key states of the \(i\)-th attention head. \cite{cai2025r} mask the elements in \(\mathbf{C}\) below a specific threshold to zeros, as well as those corresponding to the most recent tokens, resulting in a modified similarity matrix \(\mathbf{C}'\). The average similarity score for each token is computed as:
\[
\overline{\mathbf{C}}'_{i,j} = \sum_{k=0}^{l-1}{C'_{i,k,j}}.
\]
Here, \(\mathbf{C}'_{i,j}\) represents the redundancy level of the \(j\)-th token in the \(i\)-th attention head. A higher value of \(\mathbf{C}'_{i,j}\) indicates that the token is more redundant.
Finally, the redundancy score \(\mathbf R\) is obtained by applying the softmax function to the average similarity scores:
\[
\mathbf R_{i,j} = \frac{\exp(\overline{\mathbf{C}}'_{i,j})}{\sum_{k=0}^{l-1} \exp(\overline{\mathbf{C}}'_{i,k})}
\]

In Equation (\ref{eq:main}), we perform max normalization on the local scores. This approach is adopted because, as the sequence length increases, the attention distribution becomes diluted, and the average magnitude of attention scores for each token changes. 
Previous methods did not account for the combination of scores across windows, and thus normalization was unnecessary. In contrast, we mitigate this issue by applying max normalization to the local scores.
The redundancy scores calculated via the softmax function also suffer from a similar dilution problem. Therefore, when combining our global scores with the redundancy scores, we also normalize the redundancy scores as follows:  

\[
\mathbf R'_{i,j} = \frac{\mathbf R_{i,j}}{\max_j \mathbf R_{i,j}}
\]

where \(\mathbf R'_{i,j}\) represents the normalized redundancy score. Finally, we combine the global score and the redundancy score using the following formula:

\[
\mathbf F'_t = \lambda \cdot \mathbf F_t - (1 - \lambda) \cdot \mathbf R'
\]

where  \(\lambda \in [0,1]\) is a weighting factor that determines the relative contribution of the global score and the redundancy score.

\begin{figure}[h]
    \centering
    \includegraphics[width=0.8\linewidth]{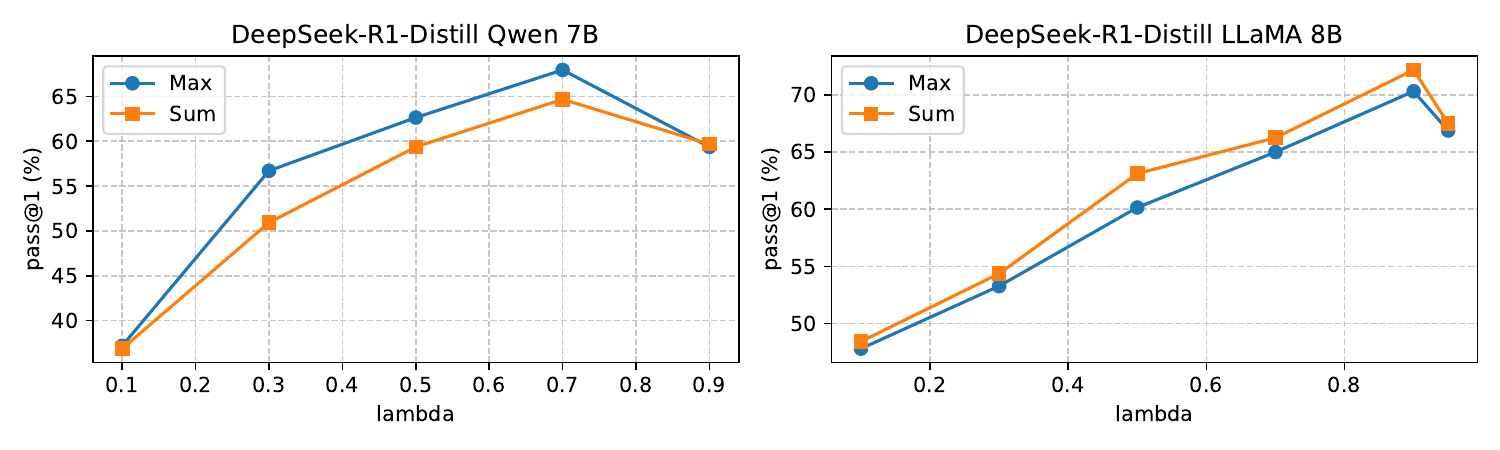}
    \caption{\textcolor{black}{Results of combining global scores with redundancy scores under different \(\lambda\) values for DeepSeek-R1-Distill Qwen-7B (left) and Llama-8B (right).}}
    \label{fig:lambda}
\end{figure}

Since we normalize the redundancy scores, we re-tune the hyperparameter \(\lambda\) instead of directly adopting the value \(\lambda=0.1\) as used in the original paper. \textcolor{black}{We fix \(\alpha\) at 0.8. Figure \ref{fig:lambda} presents the experimental results of combining global scores (max and sum) with redundancy scores under different \(\lambda\) values. For DeepSeek-R1-Distill Qwen-7B, the best performance is achieved when \(\lambda = 0.7\), while for DeepSeek-R1-Distill Llama-8B, the optimal performance is observed at \(\lambda = 0.9\). In both cases, the global score plays a dominant role.}

\section{More Information and Analysis of Training}
\label{sct:train_info}

\begin{figure}[h]
    \centering
    \includegraphics[width=0.45\linewidth]{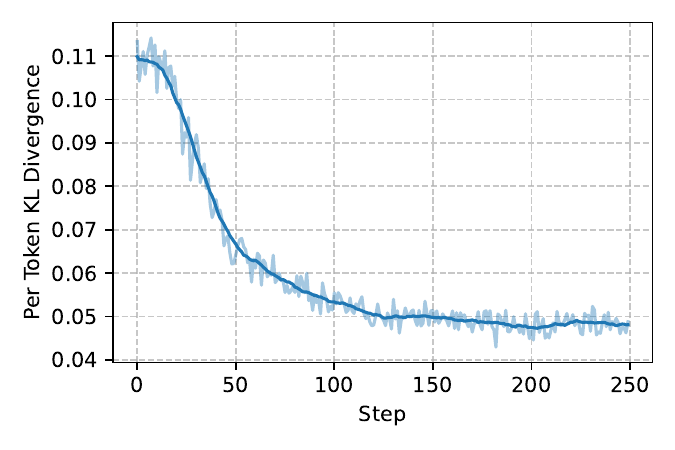}
    \caption{The average per-token KL divergence between the sparse model $\pi'\theta$ and the full attention model $\pi\theta$ during distillation training. The semi-transparent curves represent the actual values, while the solid lines indicate the smoothed values.}
    \label{fig:kl_loss}
\end{figure}

In this section, we visualize certain information recorded during the training process. Figure \ref{fig:kl_loss} illustrates the average KL divergence between the distributions of the sparse model $\pi'_\theta$ and the full attention model $\pi_\theta$ for generating the next token during distillation training. As training progresses, the KL divergence decreases rapidly, indicating that the distribution of $\pi'_\theta$ is indeed approaching that of $\pi_\theta$.

\begin{figure}[h]
    \centering
    \begin{subfigure}[t]{0.45\linewidth}
        \centering
        \includegraphics[width=\linewidth]{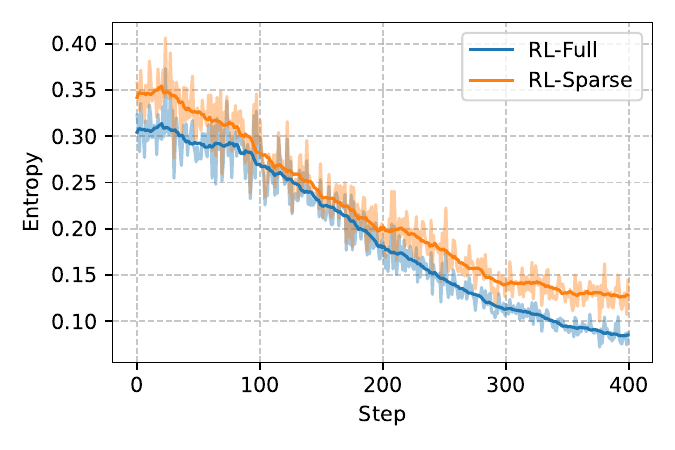}
        \caption{}
        \label{fig:entropy}
    \end{subfigure}
    \hfill
    \begin{subfigure}[t]{0.45\linewidth}
        \centering
        \includegraphics[width=\linewidth]{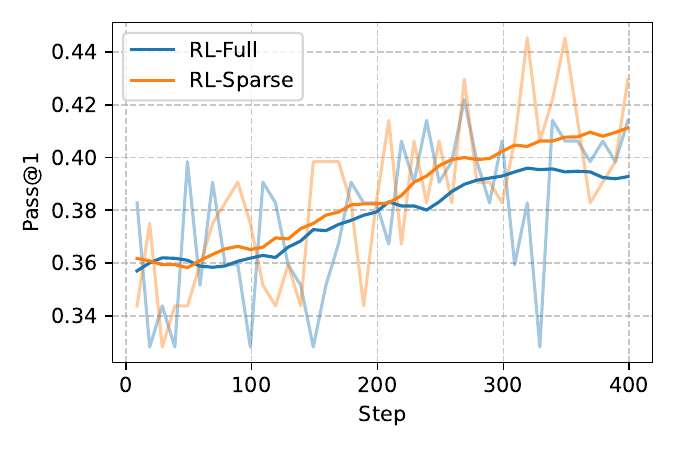}
        \caption{}
        \label{fig:eval_pass1}
    \end{subfigure}
    \caption{Changes in entropy (left) and pass@1 on the validation set (right) during reinforcement learning training. The semi-transparent curves represent the actual values, while the solid lines indicate the smoothed values. }
    \label{fig:entropy_and_pass1}
\end{figure}

Figure \ref{fig:entropy_and_pass1} (a) and (b) depict the changes in entropy \citep{shannon1948mathematical} and pass@1 on the validation set during reinforcement learning training. The validation set consists of 32 samples randomly selected from the training set. For each question in the validation set, pass@1 is estimated by sampling 4 times.

Entropy reflects the uncertainty of a distribution. The curves in Figure  \ref{fig:entropy_and_pass1} (a) indicate that RL-Sparse exhibits relatively higher entropy during the reinforcement learning process, which is due to the fact that the sparse attention mask introduces some information loss. However, as training progresses, the entropy of both RL-Sparse and RL-Full decreases rapidly. This suggests that the determinism of generation distribution from policy trained by RL-Sparse and RL-Full is improving. However, overly high determinism might lead to insufficient exploration. For larger-scale reinforcement learning, it would be beneficial to integrate advanced techniques to encourage more exploration \citep{liao2025enhancing, cui2025entropy}.

The curve of pass@1 in Figure \ref{fig:entropy_and_pass1} (b) demonstrates that RL-Sparse achieves even higher pass@1 compared to RL-Full. It is worth noting that RL-Sparse is evaluated under a compressed KV cache setting, while RL-Full is evaluated with a full KV cache. \textbf{The superior performance of RL-Sparse in terms of pass@1 may indicate the potential for faster convergence in sparse reinforcement learning.} As highlighted by \cite{wang2025beyond}, backpropagating gradients selectively on high-entropy tokens during reinforcement learning training yields better results. Sparse RL might \textbf{focus gradient updates more effectively on tokens with higher information density and greater decision-making significance, thereby improving the efficiency of policy optimization.} We believe this offers new insights for future research on reinforcement learning for LLMs.

\section{Balance between Throughput and Decoding Time}
\label{sct:time_cmp}

In \textsection \ref{sct:eff_analysis}, we compared the decoding time under the same batch size. The experimental results at that time indicated that the differences in decoding time across various budgets were minimal. This was due to the insufficient batch size, which failed to fully utilize the computational units of the GPU. In this section, we further analyze and compare the decoding times across different budgets under varying batch sizes.

\begin{figure}[h]
    \centering
    
    \begin{subfigure}[t]{0.45\linewidth}
        \centering
        \includegraphics[width=\linewidth]{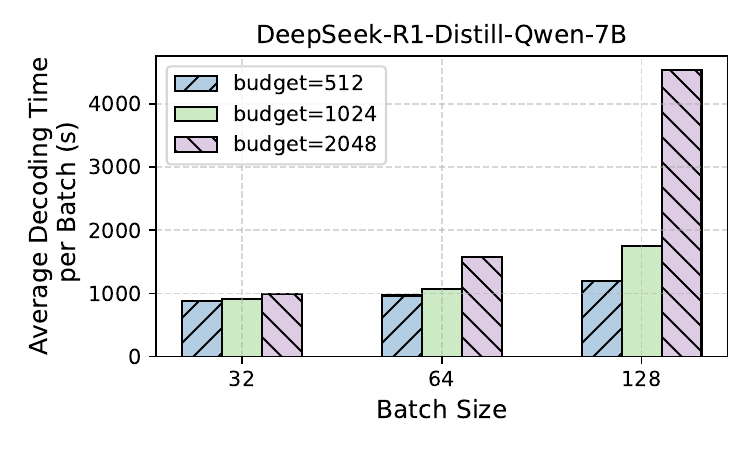}
        \label{fig:qwen_time}
    \end{subfigure}
    \hfill
    \label{fig:decoding_time_comparison}
    \begin{subfigure}[t]{0.45\linewidth}
        \centering
        \includegraphics[width=\linewidth]{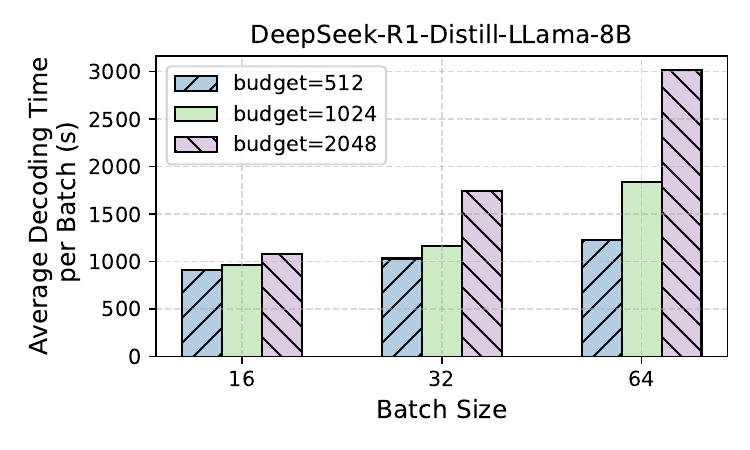}
        \label{fig:llama_time}
    \end{subfigure}
    \caption{Comparison of decoding times across different budget and batch size.}
    \label{fig:appedix_decode_time}
\end{figure}

As illustrated in Figure \ref{fig:appedix_decode_time}, an increase in batch size leads to longer decoding times. Moreover, with higher budgets, the differences in decoding time across varying batch sizes become more pronounced. For instance, when the budget is set to 512, the decoding time for a batch size of 128 is only marginally greater than that for a batch size of 32. However, when the budget increases to 2048, the decoding time surpasses more than twice the initial value. Therefore, while KV cache compression supports larger batch sizes and larger batch sizes typically yield higher throughput. However, To avoid excessively long decoding times, batch sizes should not be set excessively large.

\section{Memory Analysis}
\label{sct:memory_analysis}
\textbf{KV Cache Memory Analysis.} We take the DeepSeek-R1-Distill-Qwen-7B model as an example, which has 28 layers, an attention head dimension of 128, and 2 key-value heads. Assuming a sequence length of 16,384 (16k) and use precision of bf16 (2 bytes), the memory consumption of key and value per sequence can be calculated as
$( 28 \times 128 \times 2 \times 16384 \times 2 \times 2) / 2^{30} \approx 0.44 $ GB. When the batch size is 128, the KV cache alone requires 56 GB of memory. However, when applying KV cache compression, the required KV cache memory is reduced to $\frac{b+s}{\text{sequence length}}$,
where \( b \) denotes the KV cache budget and $s$ is the interval between two consecutive compression operations. For a batch size of 128, $s=128$ and budgets of 512, 1024, and 2048, the KV cache memory requirements are reduced to 2.18 GB, 3.93 GB, and 7.43 GB, respectively. This results in memory savings of 96.1\%, 92.9\%, and 86.7\%, respectively.

\textbf{Score Cache Memory Analysis.} Our method stores the scores computed for compression, denoted as \( \mathbf{F} \in \mathbb{R}^{h_\text{kv} \times (b-w)} \), while the shape of the key or values status cache is \( \mathbf{K}, \mathbf{V} \in \mathbb{R}^{h_\text{kv} \times b \times d} \). The memory consumption of the scores, as a fraction of the KV cache memory, is \( \frac{b-w}{b \times d \times 2} \), where \( w \ll b \). This simplifies to approximately \( \frac{1}{2 \times d} \). Since \( d \) is typically 128, the additional overhead from storing the scores is negligible compared to the memory savings achieved. \textcolor{black}{Under the same settings as in the example above, with a fixed KV cache budget of 2048, the KV cache size is approximately 7 GB, while the global score cache occupies around 27 MB.}

\textbf{Sparse Attention Mask Memory Analysis.} The memory consumption of sparse attention masks is easily overlooked; however, in practice, it can surpass even the size of the model parameters. Consider a scenario where the training batch size per device (GPU) is 16, the model consists of 28 layers, $h_\text{kv} = 2$, the sequence length is 4096 (4k), and the data type of sparse attention mask occupies only 1 byte. The memory required for the sparse attention mask is calculated as $(16 \times 28 \times 2 \times 4096 \times 4096) / 2^{30} = 14\ \text{GB}$. Loading the complete sparse attention mask onto the GPU may lead to out-of-memory (OOM) errors. To address this, the sparse attention mask is offloaded to the CPU after its construction. Furthermore, during training, \textbf{we employ gradient checkpointing and ensure that only the sparse mask for a single layer is loaded onto the GPU at any given time}. This strategy is critical for enabling training with sparse attention masks.

\section{\textcolor{black}{Experimental Results on DeepSeek-R1-Distill-LLaMA-8B}}

\begin{figure}[h]
    \centering
    \includegraphics[width=0.95 \linewidth]{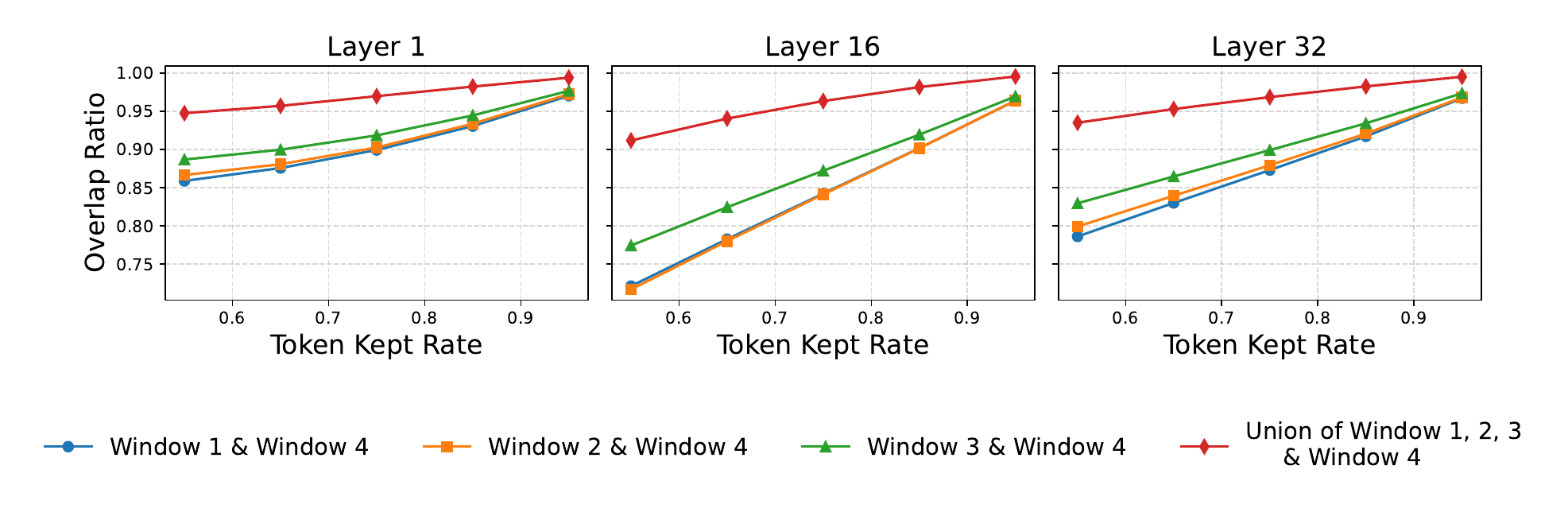}
    \caption{This figure illustrates the overlap between the set of tokens attended to in the last window and the sets of tokens attended to in other windows. The horizontal axis represents the proportion of tokens retained.}
    \label{fig:window_llama}
\end{figure}

\begin{figure}[h]
    \centering
    
    \includegraphics[width=\linewidth]{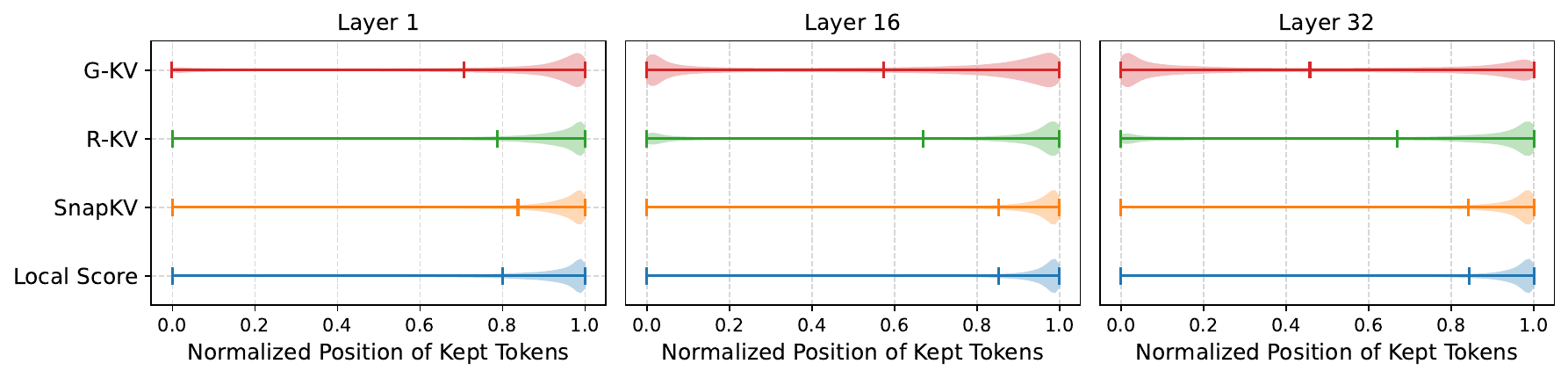}
    \caption{The density distribution of the normalized final retained token positions for different algorithms using DeepSeek-R1-Distill-LLaMa-8B. The results are evaluated on the AMC 23 benchmark. The vertical bars in the figure indicate the \textbf{mean values}.}
    \label{fig:llama_pos}
\end{figure}

Although \textsection \ref{sct:observation} and \textsection \ref{sct:main_res} only present the overlap analysis and the normalized positional density map of retained tokens for the DeepSeek-R1-Distill-Qwen-7B model, similar phenomena are observed for the DeepSeek-R1-Distill-LLaMa-8B model. The corresponding experimental results are shown in Figure \ref{fig:window_llama} and Figure \ref{fig:llama_pos}.

\begin{figure}[!h]
    \centering
    \includegraphics[width=0.65 \linewidth]{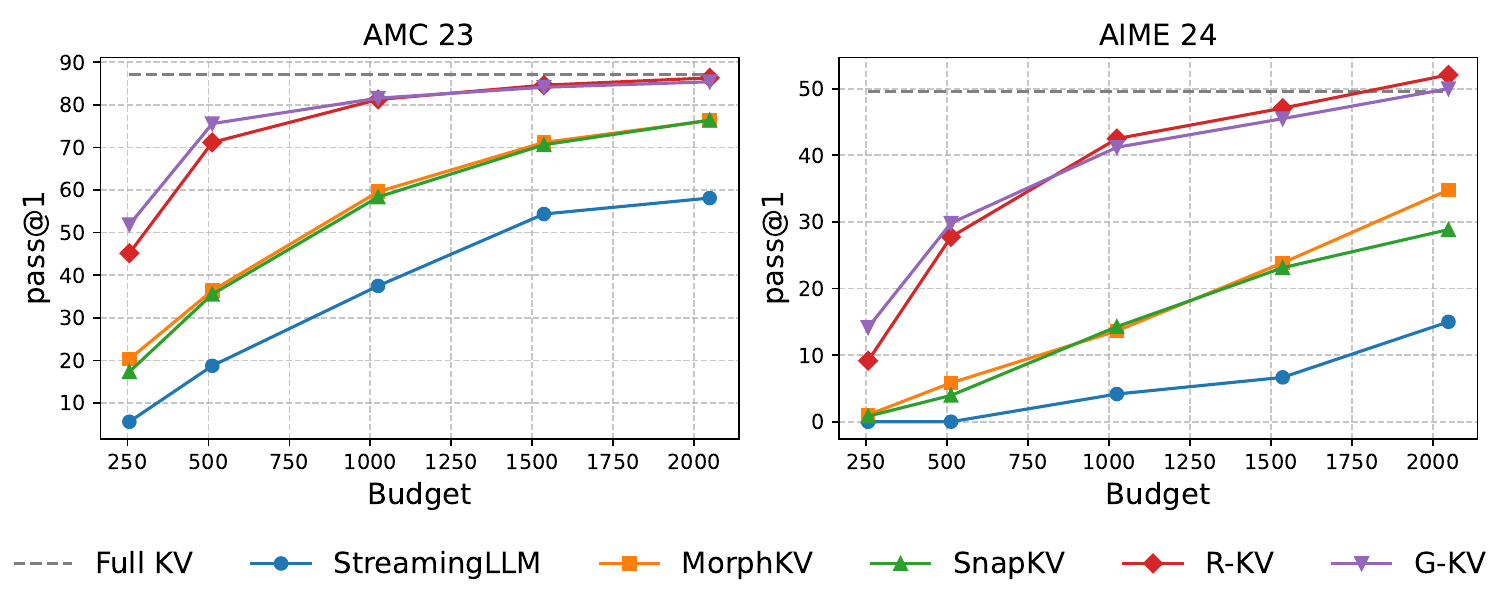}
    \caption{\textcolor{black}{Performance of different compression methods with DeepSeek-R1-Distill LLaMA 8B model.}}
    \label{fig:llama_pass1}
\end{figure}

\begin{figure}[!h]
    \centering
    \includegraphics[width=0.65 \linewidth]{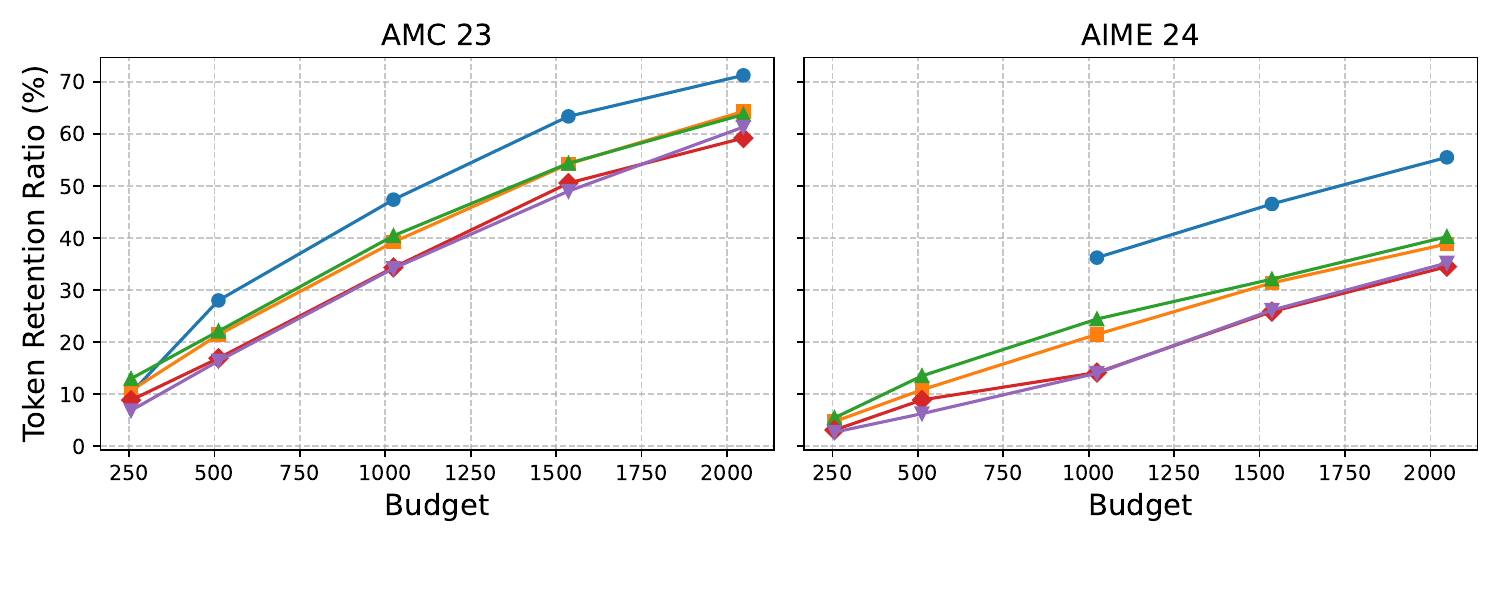}
    \caption{\textcolor{black}{Token retention ratio of different compression methods with DeepSeek-R1-Distill LLaMA 8B model.}}
    \label{fig:llama_ratio}
\end{figure}

\textcolor{black}{Figures \ref{fig:llama_pass1} and \ref{fig:llama_ratio} present the experimental results of DeepSeek-R1-Distill Llama-8B. When the budget is sufficient, both R-KV and G-KV achieve results comparable to or even surpassing those of Full KV. Under lower budgets, G-KV exhibits a certain advantage.}

\section{Discussion and Insights}

\begin{figure}[h]
    \centering
    
    \begin{subfigure}[b]{0.48\linewidth}
        \centering
        \includegraphics[width=\linewidth, trim=80 30 60 30, clip]{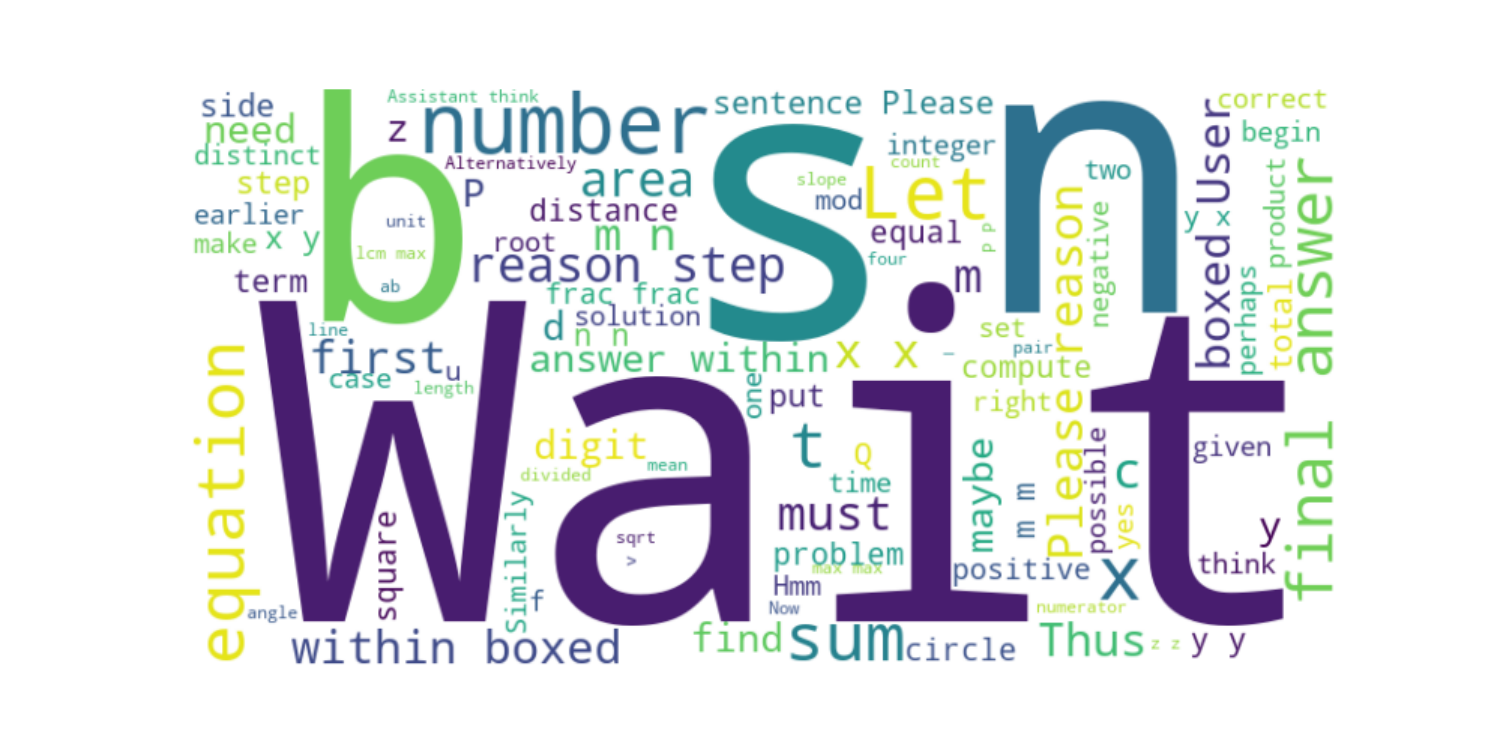}
        \caption{DeepSeek-R1-Distill-Qwen-7B}
        \label{fig:wordcloud_qwen}
    \end{subfigure}
    \hfill
    \begin{subfigure}[b]{0.48\linewidth}
        \centering
        \includegraphics[width=\linewidth, trim=80 30 60 30, clip]{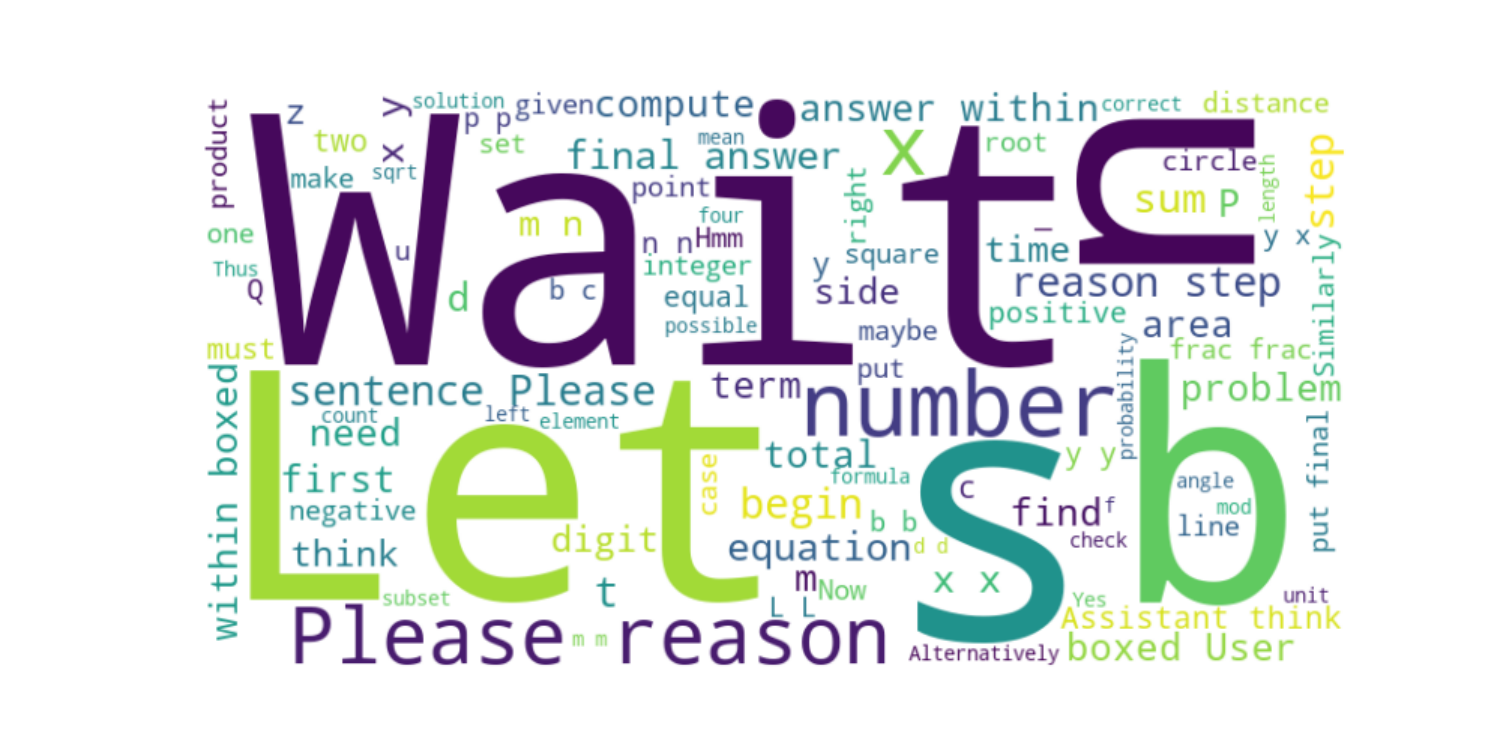}
        \caption{DeepSeek-R1-Distill-LLaMa-8B}
        \label{fig:wordcloud_llama}
    \end{subfigure}
    \caption{Word Clouds of the kept token on AMC 23}
    \label{fig:wordcloud_combined}
\end{figure}

In this section, we first visualize the tokens retained by a single KV head in the final layer using word clouds. The visualization results are presented in Figure \ref{fig:wordcloud_combined}. Notably, the word  "wait"  appears most prominently in both models. This word typically appears when the model begins to engage in reflection. \cite{shah2025rethinking} and \cite{muennighoff2025s1} have found that inserting "wait" into the output can significantly enhance the triggering of explicit reflection and improve the final accuracy.

Interestingly, combining the cases presented in Appendix \ref{sct:case_study}, we observe that the model's attention is not primarily focused on the reflective content following the word "wait," but instead is remarkably concentrated on the word "wait" itself. This phenomenon suggests that the key and value states corresponding to "wait" may have already encoded the forthcoming reflective information in advance. \textbf{In other words, the actual "thinking" process is likely to occur during the compression of information when the model accesses the value states of "wait" through the attention mechanism.}

This implies that "wait" in the deep hidden representations of LLMs carries a sufficiently high semantic density, and the subsequent reflective output merely serves to externalize the content that has already been "pre-thought" within "wait."
Of course, the processes of information compression and decoding are not confined only to "wait" but rather constitute a dynamic process throughout the decoding stage. Other tokens (e.g., periods, contrastive conjunctions, etc.) exhibit similar functionalities in deeper layers \citep{chensepllm}: through their representations, these tokens trigger the model to extract, organize, or reconstruct key information.

This further explains why deep-layer attention often exhibits high sparsity: in these layers, the KV cache representations of certain critical tokens already serve as highly compressed semantic carriers. By selectively attending to these tokens, the model can effectively accomplish the contextual integration needed for inference, without exhaustively referencing every prior token. \textbf{Interestingly, this behavior mirrors fundamental characteristics of human cognition}. When processing complex information, humans typically do not distribute their attention uniformly across all available details. Instead, they selectively focus on a small number of salient cues, which act as anchors or triggers for downstream reasoning and memory retrieval. For example, in recalling a past experience, one may only need to remember a single vivid scene—such as a spoken phrase or a specific gesture—to reconstruct the broader narrative context.

This mechanism of sparse activation and efficient recall is not merely a cognitive shortcut but a defining feature of the human memory system. It underscores a key insight: \textit{compression as intelligence}. This principle highlights the profound role of semantic compression in enabling efficient reasoning and memory retrieval. \textbf{Furthermore, it strengthens the argument for designing and training models with explicit sparsity mechanisms. Such mechanisms not only enhance computational efficiency but may also promote the development of more human-like capacities for abstraction and generalization.}


\section{Case Study}
\label{sct:case_study}

In this section, we provide a case utilizing the global score ($\alpha=0.8$) and another case employing the local score. The KV cache budget is set to 512. The questions used are sourced from AMC 23, and the model employed is DeepSeek-R1-Distill-Qwen-7B. Tokens retained by a single attention head in the final layer are highlighted in \textcolor[rgb]{1.000,0.498,0.055}{orange}, while discarded tokens are indicated in \textcolor{black}{black}. Tokens that remain uncompressed after the last compression step are marked in \textcolor[rgb]{0.173,0.627,0.173}{green}.

These two cases more intuitively illustrate the insights reflected in Figure \ref{fig:qwen_pos_qwen7b} and Figure \ref{fig:llama_pos}. When utilizing the local score, the retained tokens are concentrated towards the last part of the sequence, with most of the earlier tokens being discarded. In contrast, when employing the global score, the retained tokens are more evenly distributed throughout the entire sequence. Interestingly, a significant portion of the prompt is also preserved under the global score, which may explain its work properly on longer sequences.

\begin{figure}[h]
    \centering
    \includegraphics[width=\linewidth, trim=120 130 120 130, clip]{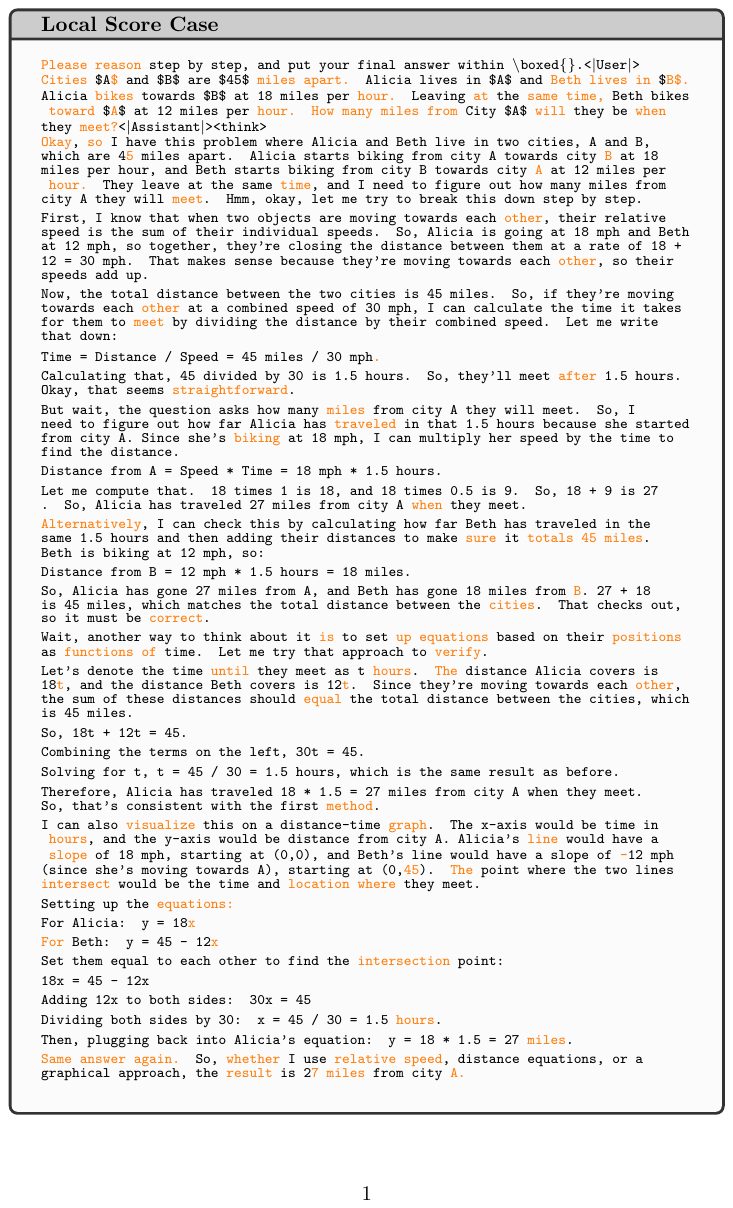}
    \caption{Part 1 of the case of local score.}
\end{figure}

\begin{figure}[h]
    \centering
    \includegraphics[width=\linewidth, trim=120 140 120 120, clip]{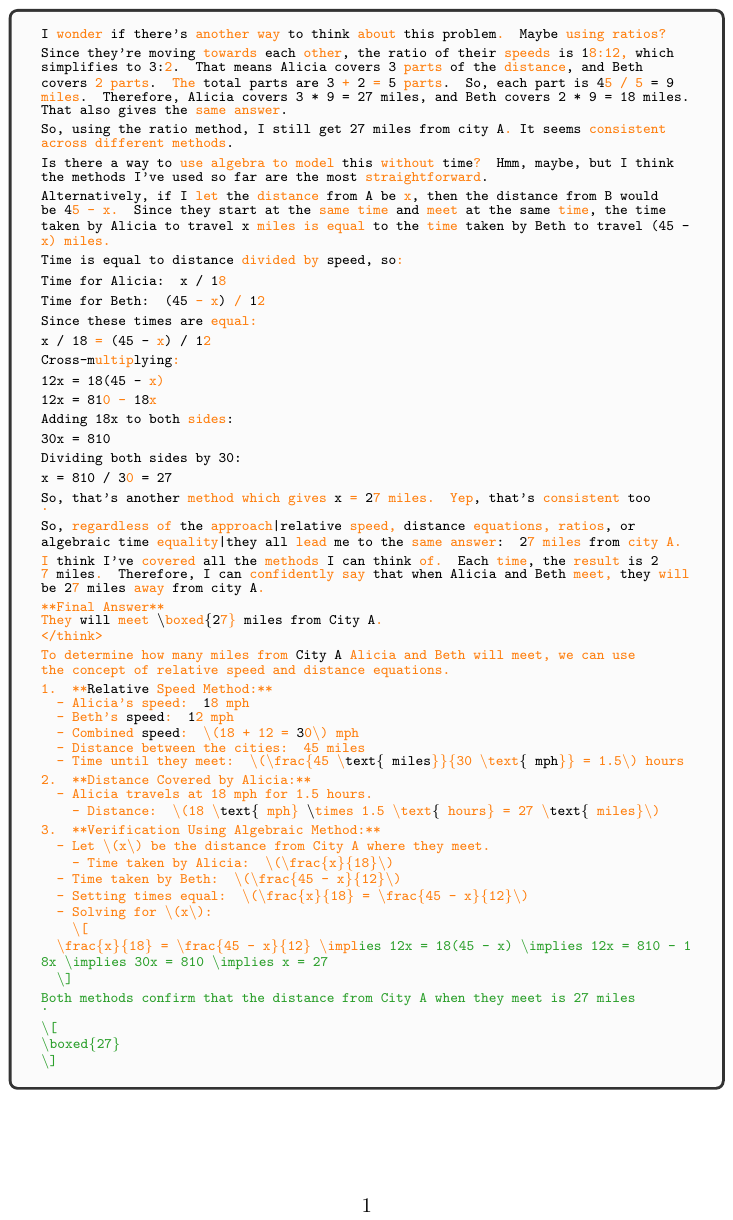}
    \caption{Part 2 of the case of local score.}
\end{figure}

\begin{figure}[h]
    \centering
    \includegraphics[width=\linewidth, trim=120 130 120 130, clip]{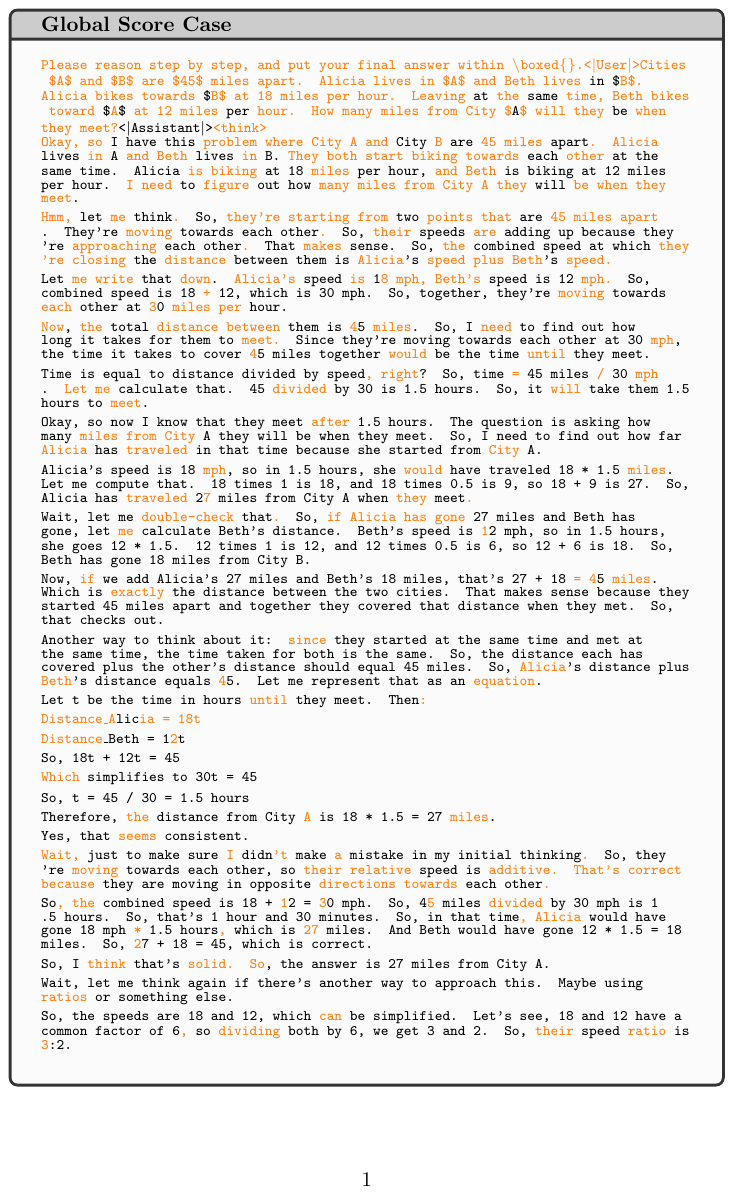}
    \caption{Part 1 of the case of global score.}
\end{figure}

\begin{figure}[h]
    \centering
    \includegraphics[width=\linewidth, trim=120 170 120 120, clip]{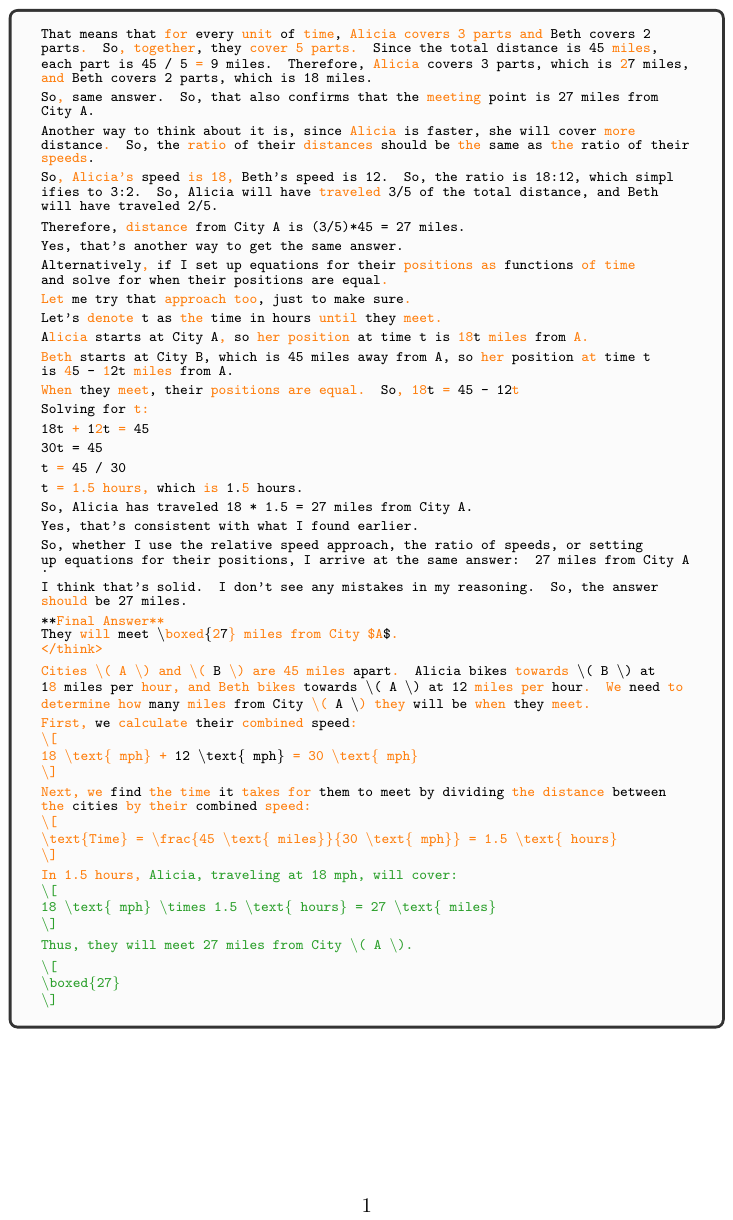}
    \caption{Part 2 of the case of global score.}
\end{figure}

\end{document}